\def\year{2021}\relax
\documentclass[letterpaper]{article} %
\usepackage{aaai21}  %
\usepackage{times}  %
\usepackage{helvet} %
\usepackage{courier}  %
\usepackage[hyphens]{url}  %
\usepackage{graphicx} %
\usepackage{graphicx}  %
\usepackage{natbib}  %
\usepackage{caption} %
\usepackage{booktabs}       %
\usepackage{multirow}       
\usepackage{amsfonts}       %
\usepackage{nicefrac}       %
\usepackage{notation} 	%
\usepackage{url} 		%
\usepackage{bm} 			%
\usepackage{kotex} 		%
\usepackage{xfrac} 		%
\usepackage{appendix} 	%
\usepackage[english]{babel} %
\usepackage{blindtext} 	%
\usepackage{soul} 		%
\usepackage[normalem]{ulem} %
\usepackage{csquotes} 	%
\usepackage{wrapfig}    %
\usepackage[inline]{enumitem}   %
\usepackage{amsmath}    %
\usepackage{caption}    %
\usepackage{subcaption} %
\usepackage{lipsum} 		%
\usepackage{todonotes}  %
\usepackage[capitalise]{cleveref}   %
\usepackage{makecell}   %
\usepackage{rotating}   %
\usepackage[switch]{lineno}  %

\title{Vector Quantized Bayesian Neural Network Inference for Data Streams}
\author{
    Namuk Park, Taekyu Lee, Songkuk Kim
}
\affiliations{

    Yonsei University

    \{namuk.park, taekyulee, songkuk\}@yonsei.ac.kr

}

\begin{document}

\maketitle

\begin{abstract}

Bayesian neural networks (BNN) can estimate the uncertainty in predictions, as opposed to non-Bayesian neural networks (NNs). However, BNNs have been far less widely used than non-Bayesian NNs in practice since they need iterative NN executions to predict a result for one data, and it gives rise to prohibitive computational cost. This computational burden is a critical problem when processing data streams with low-latency. To address this problem, we propose a novel model VQ-BNN, which approximates BNN inference for data streams. In order to reduce the computational burden, VQ-BNN inference predicts NN only once and compensates the result with previously memorized predictions. To be specific, VQ-BNN inference for data streams is given by temporal exponential smoothing of recent predictions. The computational cost of this model is almost the same as that of non-Bayesian NNs. Experiments including semantic segmentation on real-world data show that this model performs significantly faster than BNNs while estimating predictive results comparable to or superior to the results of BNNs.

\end{abstract}

\section{Introduction}

While deterministic neural networks show high accuracy in many areas, they cannot estimate reliable uncertainty. Predictions cannot be perfect and some incorrect predictions might bring about fatal consequences in areas such as medical analysis and autonomous vehicles control. Therefore, estimating uncertainty as well as predictions is crucial for the safer application of machine learning based systems.

Bayesian neural network (BNN) uses probability distributions to model neural network (NN) weights and estimates not only predictive results but also uncertainties. This allows computer systems to make better decisions by combining prediction with uncertainty. Moreover, BNNs can achieve high performance in a variety of fields, e.g. image recognition \cite{kendall2015bayesian,kendall2017uncertainties}, language modeling \cite{fortunato2017bayesian}, reinforcement learning \cite{kahn2017uncertainty,osband2018randomized}, meta-learning \cite{yoon2018bayesian,finn2018probabilistic}, and multi-task learning \cite{kendall2018multi}.

Despite these merits, BNNs have a major disadvantage that make it difficult to use as a practical tool; the predictive inference speed of BNNs is dozens of times slower than that of deterministic NNs. It has held back BNNs from wide applications. Particularly, this problem is a significant barrier for processing data streams with low-latency. This will be further elaborated below.

\paragraph{BNN inference.}

Let $p(\lats \vert \mathcal{D})$ be a posterior probability of NN weights $\lats$ with respect to training dataset $\mathcal{D}$, and $p(\outs \vert \dats, \lats)$ be a probability distribution parameterized by NN's result for an input data vector $\dats$ and a weight $\lats$. Then, the inference result of BNN is a predictive distribution:
\begin{equation}
	p(\outs \vert \dats_{0}, \mathcal{D}) = \int p(\outs \vert \dats_{0}, \lats) \, p(\lats \vert \mathcal{D}) d\lats \label{eq:bnn-inference}
\end{equation}
where $\dats_{0}$ is observed input data vector and $\outs$ is output vector. Since this equation cannot be solved analytically, we use the MC estimator to approximate it:
\begin{equation}
	p(\outs \vert \dats_{0}, \mathcal{D}) \simeq \sum_{\lats_{i}} \frac{1}{N_{\lats}} \, p(\outs \vert \dats_{0}, \lats_{i}) \label{eq:approx-bnn-inference}
\end{equation}
where $\lats_{i} \sim p(\lats \vert \mathcal{D})$ and $N_{\lats}$ is the number of the samples. 
The MC estimator implies that NN needs to be executed iteratively to calculate the predictive distribution. As many real-world data is large and practical NNs are deep, multiple NN execution cannot be fully parallelized \cite{kendall2017uncertainties}. Consequently, the computation speed is significantly decreased. For example, according to \cref{sec:bnn-performance-for-forward-pass} and \cite{kendall2017uncertainties}, BNN requires up to fifty predictions to obtain high predictive performance in computer vision tasks, which means that the data processing speed of BNN could be fifty times lower than that of deterministic NN.

\paragraph{VQ-BNN inference.}

\begin{figure*}[ht]
\vskip 0.2in
\begin{center}
\begin{subfigure}[b]{0.43\textwidth}
    \centering
    \includegraphics[width=\textwidth,page=1]{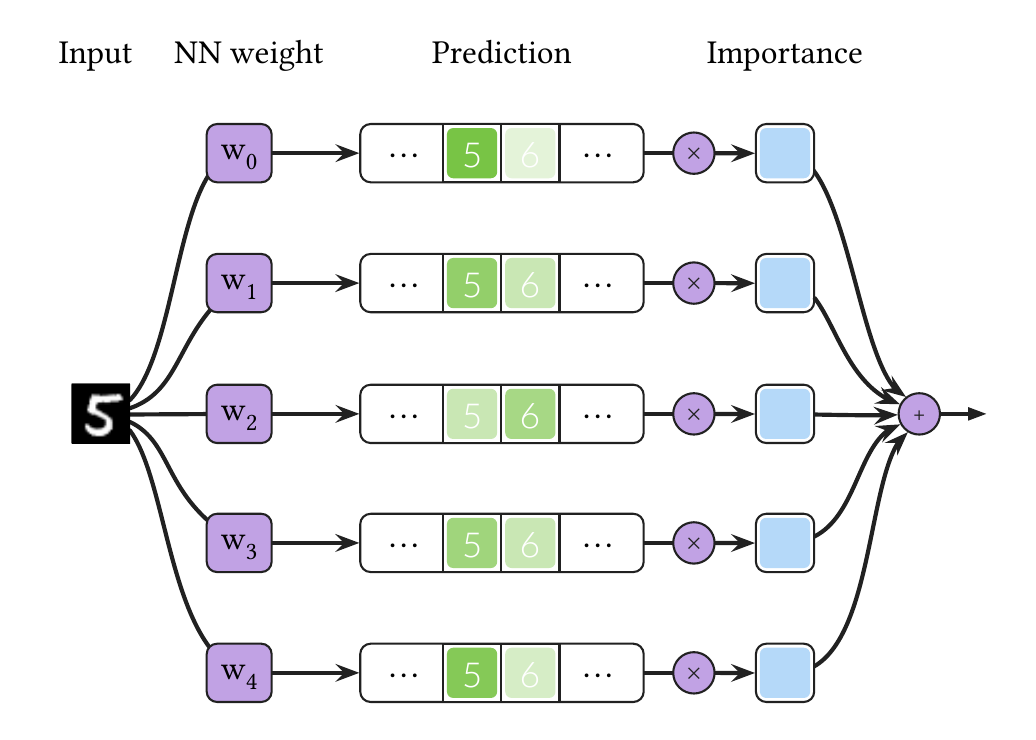}
    \caption{BNN Inference}
\end{subfigure}
\begin{subfigure}[b]{0.452\textwidth}
    \centering
    \includegraphics[width=\textwidth,page=2]{resources/diagrams} 
    \caption{VQ-BNN Inference}
\end{subfigure} 
\vskip 0.05in

\caption{
\textbf{Comparison of BNN inference and VQ-BNN inference.}
The predictive distribution of BNN inference is the sum of the probabilities $\{ p(\outs \vert \dats_{0}, \lats_{i}) \}$ parameterized by NN's results---e.g. for classification tasks, $p(\outs \vert \dats_{0}, \lats_{i}) = \texttt{Softmax}(\nn (\dats_{0}, \lats_{i}))$ where $\nn(\cdot)$ is logit of NN---for the same observed input data and different NN weights.
The predictive distribution of VQ-BNN inference is the importance weighted sum of one prediction $p(\outs \vert \dats_{0}, \lats_{0})$ for the observed data and the previously memorized predictions $\{ p(\outs \vert \dats_{i}, \lats_{i}) \}$ for different inputs and weights. The importance is defined as the similarity between the observed data and memorized data.
VQ-BNN inference for continuously changing data streams is temporal smoothing of recent predictions with exponentially decaying importances because we assume that the similarity between the latest data and the past data decreases exponentially over time. In this figure, the inputs are toy examples.
}

\label{fig:vqbnn-inference}
\end{center}
\vskip 0.0in
\end{figure*}

Suppose we have access to memorized input dataset $\{ \dats_{0}, \dats_{1}, \cdots \}$ consisting of similar data and corresponding predictions with different weights $\{ p(\outs \vert \dats_{0}, \lats_{0}), p(\outs \vert \dats_{1}, \lats_{1}), \cdots \}$ in the process of inference. Then, the predictive distribution of BNN can be approximated by combining these predictions. Based on this idea, we propose novel predictive distribution called \emph{vector quantized BNN} (VQ-BNN) inference that approximates BNN inference by using the quantized vectors to speed up calculating the predictive distribution. 

In order to reduce the computational burden, VQ-BNN inference performs NN prediction for the observed input $\dats_{0}$ only once. Then, it compensates the result with previously memorized predictions. We expect that the predictive distribution of VQ-BNN is analogous to that of BNN, since NN produces similar predictions for similar inputs.
For a more sophisticated approximation, the importance of the prediction in the predictive distribution is determined based on the similarity between the observed input $\dats_{0}$ and the prediction's input $\dats_{i}$.
To sum up, VQ-BNN inference is as follows:
\begin{equation}
	p(\outs \vert \dats_{0}, \mathcal{D}) \simeq \sum_{(\dats_{i}, \lats_{i})} \pi(\dats_{i} \vert \dats_{0}) \, p(\outs \vert \dats_{i}, \lats_{i})
	\label{eq:approx-vqbnn-inference-1}
\end{equation}
where $\lats_{i} \sim p(\lats \vert \mathcal{D})$ and $\pi(\dats_{i} \vert \dats_{0})$ is an importance of $\dats_{i}$ with respect to $\dats_{0}$. To estimate this predictive distribution, only $p(\outs \vert \dats_{0}, \lats_{0})$ needs to be calculated, since remainders  are obtained from memorized predictions.
This makes the computational performance of VQ-BNN comparable to that of deterministic NN.

VQ-BNN inference requires memorizing input vectors, similar to the observed input data $\dats_{0}$, and corresponding predictions. To obtain them, we suppose that most of the time-varying data streams are continuously changing. Based thereupon, we prepare the proximate data sequence for VQ-BNN inference on data streams by memorizing the last few data and NN predictions. 
Also, we propose the importance of a previous data that decreases exponentially over time, i.e., $\pi(\dats_{i} \vert \dats_{0}) = \exp(\sfrac{- \Delta t_i}{\tau}) / \sum_{i} \exp(\sfrac{- \Delta t_i}{\tau})$ where $\tau$ is hyperparameter and $\Delta t_i$ is the time difference between $\dats_{i}$ and $\dats_{0}$.
In conclusion, VQ-BNN inference for data streams is \emph{temporal smoothing} with exponentially decaying importance of recent predictions. We summarize VQ-BNN inference in \Cref{fig:vqbnn-inference}.

\paragraph{Results.}

We evaluate VQ-BNN with computer vision tasks namely semantic segmentation and depth estimation on a variety of high-dimensional video sequence datasets. The results show that VQ-BNN has almost no degradation in computational performance compared to deterministic NNs. The predictive performance of VQ-BNN is comparable to or superior to that of BNN in various situations.

\paragraph{Contributions.}

The main contributions of this work are as follows.

\begin{itemize}
	\item We propose vector quantized Bayesian neural network (VQ-BNN) inference as an approximation of Bayesian neural network inference to enhance the computational performance.
	\item We propose temporal smoothing of predictions with exponentially decaying importance by applying VQ-BNN inference to data streams.
	\item We empirically show that the computational performance of VQ-BNN is almost the same as that of deterministic NN and the predictive performance is comparable to or better than that of BNN on real-world data streams.
\end{itemize}

\vskip 0.0in

\section{Vector Quantized \\Bayesian Neural Network Inference} \label{sec:vqbnn}

Let $\strs$ be a set of data points $\{ \dats_{0}, \cdots, \dats_{K} \}$ generated by a source and $p(\dats \vert \strs)$ be an estimated probability distribution of the set of data. The data points are also known as \emph{prototypes} because they represent the probability. When the source is stationary, the estimated probability represents the observation noise.

We propose a predictive distribution for $\strs$ as an alternative to the predictive distribution of BNN for one data point $\dats_{0}$:
\begin{align}
	p(\outs \vert \strs, \mathcal{D}) &= \int p(\outs \vert \dats, \lats) \, p(\dats \vert \strs) \, p(\lats \vert \mathcal{D}) \, d\dats d\lats \\
	&= \int p(\outs \vert \inps) \, p(\inps \vert \strs, \mathcal{D}) \, d\inps 
	\label{eq:vqbnn-inference}
\end{align}
For simplicity, we introduce $\inps = (\dats, \lats)$ and $p(\inps \vert \strs, \mathcal{D}) = p(\dats \vert \strs) \, p(\lats \vert \mathcal{D})$ in this expression. 
We call $p(\dats \vert \strs)$ \emph{data uncertainty} and $p(\lats \vert \mathcal{D})$ \emph{model uncertainty}.

In general, \cref{eq:vqbnn-inference} cannot be solved analytically. We obtain \emph{VQ-BNN inference}, i.e., 
\begin{align}
	p(\outs \vert \strs, \mathcal{D}) \simeq \sum_{\inps_{i}} \pi(\dats_{i} \vert \strs) \, p(\outs \vert \inps_{i}) 
	\label{eq:approx-vqbnn-inference}
\end{align}
by approximating $p(\inps \vert \strs, \mathcal{D})$. In this equation, we use the following quantized vector samples with importances: 
\begin{align}
	\left(\inps_{i}, \pi(\inps_{i} \vert \strs, \mathcal{D})\right) \sim p(\inps \vert \strs, \mathcal{D})
	\label{eq:importance-sampling}
\end{align}
where $\inps_{i}$ is a joint of a prototype $\dats_{i} \in \strs$ and a random NN weight sample $\lats_{i} \sim p(\lats \vert \mathcal{D})$, i.e., $\inps_{i} = (\dats_{i}, \lats_{i})$. Then, $p(\outs \vert \inps_{i})$ is a NN prediction for $\dats_{i}$ with a random weight.
$\pi(\inps_{i} \vert \strs, \mathcal{D})$ is the importance of $\inps_{i}$ with $\sum_{i=0}^{K} \pi(\inps_{i} \vert \strs, \mathcal{D}) = 1$. 
In \cref{eq:approx-vqbnn-inference}, we assume that $\pi(\inps_{i} \vert \strs, \mathcal{D}) \simeq \pi(\dats_{i} \vert \strs)$ because $\lats_{i}$ is i.i.d.. 
VQ-BNN inference implies that importances and predictions are required to obtain the predictive distribution.
\Cref{eq:approx-vqbnn-inference} is equivalent to \cref{eq:approx-vqbnn-inference-1} except that the set of prototypes is denoted by $\dats_{0}$ instead of $\strs$.

Consider the case where the prototypes are given from a noiseless stationary source, i.e., $p(\dats \vert \strs) = \delta(\dats-\dats_{0})$. In this case, all prototypes and importances are the same, and all predictions $p(\outs \vert \dats_{i}, \lats_{i})$ become $p(\outs \vert \dats_{0}, \lats_{i})$. As a result, VQ-BNN inference which is \cref{eq:approx-vqbnn-inference} reduces to BNN inference which is \cref{eq:approx-bnn-inference}. In the same manner, when $\strs$ consists of data proximate to $\dats_{0}$, the predictive distribution of VQ-BNN is similar to that of BNN.

We can improve the computational performance of calculating predictive distribution by using VQ-BNN inference. 
Without loss of generality, let $\dats_{0}$ be the observed input data. Also, suppose that we have access to memorized prototypes $\{ \dats_{1}, \cdots, \dats_{K} \}$ and the corresponding predictions $\{ p(\outs \vert \inps_{1}), \cdots, p(\outs \vert \inps_{K}) \}$.
To calculate the predictive distribution of VQ-BNN, $\pi(\dats_{i} \vert \mathcal{S})$ for all prototypes and only $p(\outs \vert \inps_{0})$ for $\inps_{0}$ are required since the remainders are obtained from the memorized predictions.
Because the time to calculate importances and to aggregate memorized predictions are negligible, 
it takes almost the same amount of time to perform VQ-BNN inference and to perform NN prediction once.

\paragraph{The case of data stream.}

In order to use VQ-BNN as the approximation theory of BNN, we have to take the proximate dataset as prototypes and derive the importances of the prototypes. To calculate the predictive distribution for data streams, VQ-BNN exploits the fact that most real-world data streams change continuously.

Thanks to the temporal proximity of data stream, we take the latest data and the recent subsequence from the data stream as proximate prototypes as follows:
\begin{align}
	\strs = \{ \dats_{t} \vert \, 0 \geq t \geq - K \}
	\label{eq:prototype}
\end{align}
where $t$ is integer timestamp and $K$ is non-negative number of prototypes from old data streams. In most cases, we need to derive the NN predictions for every data points from the data stream, and it is easy to memorize the sequence of NN predictions $\{ p(\outs \vert \inps_{t}) \}$.

We define the importance in a similar way as above. Temporal proximity of data stream implies that older data contributes less to the estimated probability distribution for a data stream. Based on this idea, we propose a model in which the importance decreases exponentially over time as follows:
\begin{align}
	\pi(\dats_{t} \vert \strs) = \frac{\exp(-\sfrac{\vert t \vert}{\tau})}{\sum_{t=0}^{-K} \exp(\sfrac{-\vert t \vert}{\tau})}
	\label{eq:exponential-importance}
\end{align}
where $\tau$ is given non-negative parameter. $\tau$ is determined experimentally depending on the characteristics of model and data stream. As $\tau$ approaches $0$, the latest prototypes will mainly contribute to the results. As $\tau$ approaches $\infty$, old prototypes also will equally contribute to the results. In summary, VQ-BNN inference for data stream is temporal smoothing of recent predictions of BNN with exponentially decaying importances.

We have to mention that if the input vector is high-dimensional, VQ-BNN might need a very large number of prototypes to represent the probability of a dataset. The more prototypes are required, the more memory is required, which makes VQ-BNN inference impractical. Despite these concerns, VQ-BNN achieve high prediction performance by using a very small number of prototypes. This is because the relevant data in the data stream are concentrated in a short time interval. \cref{app:hyperparameters} shows that the semantic segmentation task on a real-world video sequence requires the recent 5 frames to obtain high predictive performance.

There are more complex algorithms that can be used to estimate prototypes from a data stream. For example, \cite{xu2012incremental, frezza2014online, ghesmoune2016new} proposed algorithms that change prototypes depending on data stream. 
However, these algorithms are not suitable for VQ-BNN since they are too complicated and slow. It is an important drawback because VQ-BNN is developed to achieve high computational performance to process data. The experimental results shows that this simple importance model can achieve high predictive performance.

\paragraph{Implementation. }

In order to calculate VQ-BNN inference, we have to determine the prediction $p(\outs \vert \inps_{i})$ parameterized by NN. For classification tasks, we set $p(\outs \vert \inps_{i})$ as a categorical distribution parameterized by the softmax of NN logit:
\begin{align}
	p(\outs \vert \strs, \mathcal{D}) \simeq \sum_{t=0}^{-K} \pi(\dats_{t} \vert \strs) \, \texttt{Softmax}(\nn (\dats_{t}, \lats_{t}))
	\label{eq:vqbnn-classification}
\end{align}
where $\nn(\cdot)$ is logit of NN, $\dats_{t}$ is given by \cref{eq:prototype}, $\lats_{t} \sim p(\lats \vert \mathcal{D})$, and $\pi(\dats_{t} \vert \strs)$ is given by \cref{eq:exponential-importance}. For regression tasks, $p(\outs \vert \inps_{i})$ are usually modeled to have a Gaussian distribution with the mean of the NN's result:
\begin{align}
	p(\outs \vert \strs, \mathcal{D}) \simeq \sum_{t=0}^{-K} \pi(\dats_{t} \vert \strs) \, \mathcal{N}(\outs \vert \nn (\dats_{t}, \lats_{t}), \sigma^{2})
	\label{eq:vqbnn-regression}
\end{align}
where $\sigma$ is a given parameter.

For stream processing, we further simplify the VQ-BNN inference with exponentially decaying importance. Let $q_{t^{\prime}}(\outs \vert \strs, \mathcal{D})$ be the predictive distribution for prototypes in $t^{\prime} \geq t \geq -\infty$. Then, we rewrite $q_{0}(\outs \vert \strs, \mathcal{D})$ as follows:
\begin{align}
	q_{0}(\outs \vert \strs, \mathcal{D}) 
	&=  \sum_{t=0}^{-\infty} \alpha \exp(\sfrac{- \vert t \vert}{\tau}) \, p(\outs \vert \inps_{t}) \\
	&= \alpha \, p(\outs \vert \inps_{0}) + (1 - \alpha) \, q_{-1}(\outs \vert \strs, \mathcal{D})
\end{align}
where $\alpha = \left({\sum_{t=0}^{-\infty} \exp(\sfrac{- \vert t \vert}{\tau})} \right)^{-1}$. 
According to this equation, the predictive distribution is the mixture of the latest prediction and the previous predictive distribution.

\paragraph{Training. }

The loss function of a BNN, such as evidence lower bound (ELBO) or negative log-likelihood (NLL), depends on a predictive distribution. Therefore, we can calculate the loss function by using VQ-BNN inference instead of by using BNN inference when training NNs. 

However, \emph{we obtain the posterior distribution in the same way as BNN training} for some practical limitations. 
First, VQ-BNN inference depends on the order of the input data stream. It increases the implementation complexity of training with VQ-BNN inference. 
Next, many training datasets do not have all the labels corresponding to the input data stream. To derive the predictive distribution for an input with a label, VQ-BNNs have to predict the result for the previous inputs without a label. It significantly increases the time required for the NN training process.
Experiments show that VQ-BNN inference achieves high predictive performance even though it uses the posterior distribution by BNN training.

\section{Experiments}\label{sec:experiment}

\newcommand{\imgsize}[0]{0.15}
\newcommand{\skipsize}[0]{0.03}

\begin{figure}
     \centering
     \begin{subfigure}[b]{\imgsize\textwidth}
         \centering
         \includegraphics[width=\textwidth]{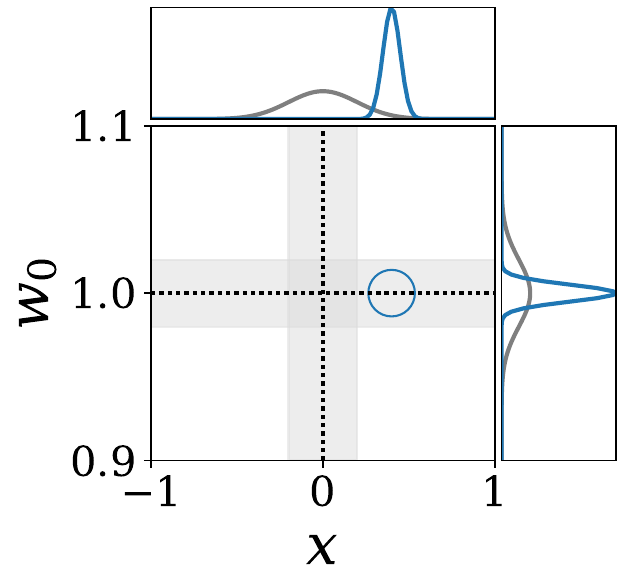}
		 \vskip \skipsize in
         \includegraphics[width=\textwidth]{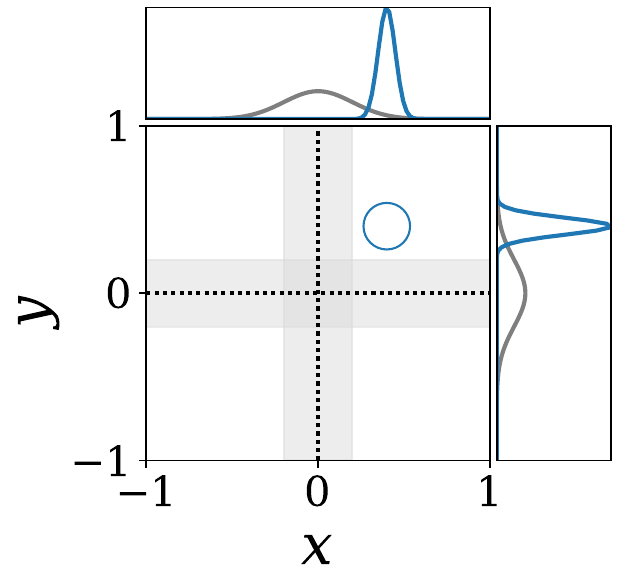}
         \caption{DNN}
         \label{fig:simple-linear-regression:dnn}
     \end{subfigure}
     \begin{subfigure}[b]{\imgsize\textwidth}
         \centering
         \includegraphics[width=\textwidth]{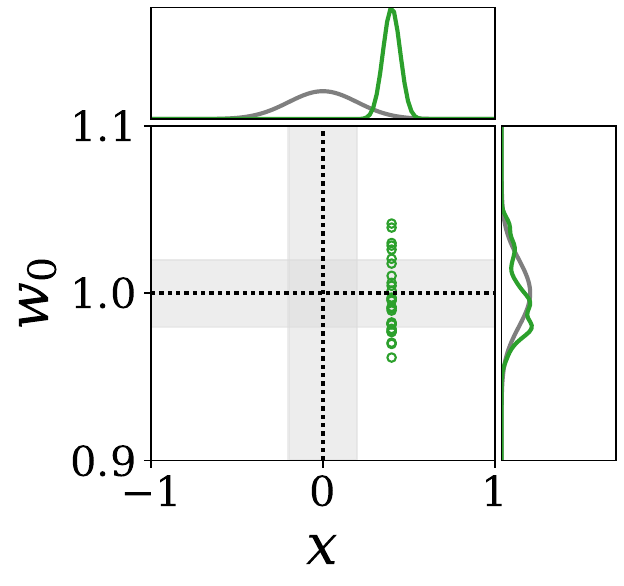}
		 \vskip \skipsize in
         \includegraphics[width=\textwidth]{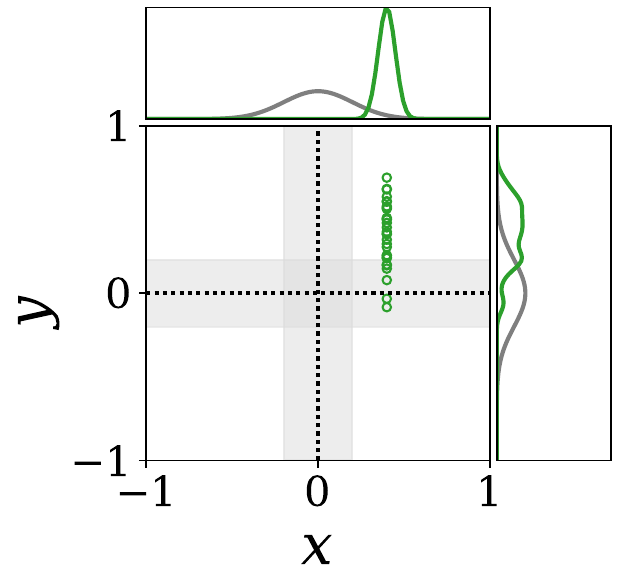}
         \caption{BNN}
         \label{fig:simple-linear-regression:mu}
     \end{subfigure}
     \begin{subfigure}[b]{\imgsize\textwidth}
         \centering
         \includegraphics[width=\textwidth]{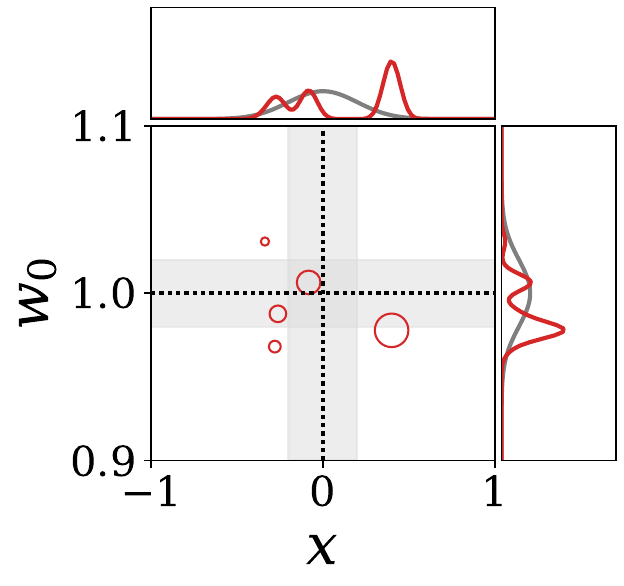}
		 \vskip \skipsize in
         \includegraphics[width=\textwidth]{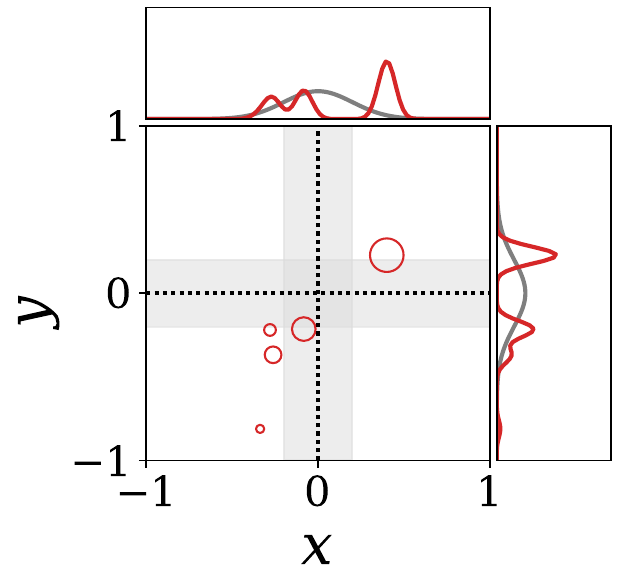}
         \caption{VQ-BNN}
         \label{fig:simple-linear-regression:vqbnn}
     \end{subfigure}
     
     \caption{
     \textbf{Visualization of VQ-BNN} with simple linear regression.
     The top are approximated distributions of input and NN weight prototypes $p(\datsimples, \latsimples_{0} \vert \strs, \mathcal{D})$ and the bottom are approximated distributions of output prototypes with data $p(\datsimples, \outsimples \vert \strs, \mathcal{D})$ at $t = 0$. 
     The sizes of the circles indicate the importances of each prototype. 
     They also show marginal distributions $p(\datsimples \vert \strs)$, $p(\latsimples_{0} \vert \mathcal{D})$, and $p(\outsimples \vert \strs, \mathcal{D})$.
     In \Cref{fig:simple-linear-regression:vqbnn}, data points at $\datsimples < 0$ are memorized prototypes from the past data stream.
     The black dotted lines and gray distributions represent true values. The error is $80\%$ confidence interval.
     }
     \label{fig:simple-linear-regression}
\end{figure}

\begin{table*}[t]
\vskip 0.1in
\begin{center}
\begin{small}
\begin{sc}

  \begin{tabular}{lccccccccccc}
    \toprule
    Method & \thead{Bat Thr\\(Img/Sec)} & \thead{Str Thr\\(Img/Sec)} & NLL & \thead{Acc\\(\%)} & \thead{Acc$_{90}$\\(\%)} & \thead{Unc$_{90}$\\(\%)} & \thead{IoU\\(\%)} & \thead{IoU$_{90}$\\(\%)} & \thead{Freq$_{90}$\\(\%)} & \thead{ECE\\(\%)} \\
    \midrule
    DNN    & \textbf{27.5} & \textbf{10.5} & 0.314 & 91.1 & 96.1 & 61.3 & 66.1 & 77.7 & 86.4 & 4.31 \\
    BNN    & 0.824 & 0.788 & 0.276 & 91.8 & 96.5 & 63.0 & 68.1 & 79.9 & \textbf{86.8} & 3.71 \\
    VQ-BNN & 25.5 & 9.41 & \textbf{0.253} & \textbf{92.0} & \textbf{97.4} & \textbf{72.4} & \textbf{68.6} & \textbf{83.7} & 83.1 & \textbf{2.24} \\

    \bottomrule
  \end{tabular}

\end{sc}
\end{small}
\end{center}
\vskip -0.08in
\caption{
\textbf{Computational and predictive performance with semantic segmentation} for each method.
}
\label{tab:semantic-segmentation-performance}
\end{table*}

This section evaluates the performance of VQ-BNN in three sets of experiments. The first experiment visualizes the characteristics of VQ-BNN with simple linear regression on synthetic data. 
The second experiment performs semantic segmentation on high-dimensional real-world video sequences. This classification task compares the performances of VQ-BNN with other baselines of deep NNs in a practical situation. 
The last experiment performs monocular depth estimation on high-dimensional real-world video sequences. This experiment compares the performance of VQ-BNN in a regression task.

\paragraph{Baselines. }

We compare the following three methods in the experiments:
\begin{itemize}
	\item \textbf{DNN. } 
	Let $\texttt{Softmax}(\hat{\outs})$ be predictive probability of deterministic NN (DNN) where $\hat{\outs}$ is NN logits for classification tasks. It is easy to implement, but it deviates from the true classification probability when the NN is deepened, broadened, and regularized well \cite{guo2017calibration}.
	\item \textbf{BNN. }
	BNNs use the MC estimator \cref{eq:approx-bnn-inference} to calculate a predictive distribution. It is difficult to analytically determine the sufficient number of NN weight samples to converge predictive distribution. Instead, we experimentally set the number of the samples to 30---i.e., BNNs with MC dropout \cite{Gal:2016uw} layers predict results with 30 forward passes in \cref{sec:semantic-segmentation} and \cref{sec:depth-estimation}---so that the negative log-likelihood (NLL) converge. \Cref{sec:bnn-performance-for-forward-pass} shows the predictive performance of BNN for different numbers of forward passes. 
	\item \textbf{VQ-BNN. } 
	As explained in \cref{sec:vqbnn}, VQ-BNN inference uses the same model and weight distribution as BNN. In all experiments, we use the same hyperparameters, $K=5$ and $\tau=1.25$, which implies that VQ-BNN is not overly sensitive to hyperparameter selection. See \cref{app:hyperparameters} for performance changes according to hyperparameters.
\end{itemize}

\subsection{Simple Linear Regression}\label{sec:simple-linear-regression}

This experiment uses a linear regression model $\outsimples = \latsimples_{0} \datsimples + \latsimples_{1}$ to find out the characteristics of VQ-BNN. 
The posteriors for BNN and VQ-BNN are given by $p(\latsimples_{0} \vert \mathcal{D}) = \mathcal{N}(1.0, 0.02^2)$ and $p(\latsimples_{1} \vert \mathcal{D}) = \mathcal{N}(0.0, 0.2^2)$. The weights for DNN are expected values of the posteriors, i.e., $\latsimples_{0} = 1.0$ and $\latsimples_{1} = 0.0$.
The distribution of time-varying input data streams is given by $p(\datsimples \vert t) = \mathcal{N}(\datsimples \vert v t, 0.1^2)$ where $t$ is integer timestamp from $-10$ and $v = 0.01$.

\paragraph{Results. }

\Cref{fig:simple-linear-regression} shows the probability distributions approximated by prototypes at $t = 0$. In this figure, the upper row displays approximated distributions of input and NN weight prototypes, i.e., $p(\datsimples, \latsimples_{0} \vert \strs, \mathcal{D})$, and the lower row shows approximated distributions of output prototypes with data, i.e., $p(\datsimples, \outsimples \vert \strs, \mathcal{D})$. The sizes of the circle indicate the importances of each prototype.
It also show the three kinds of marginal distributions: the probability distribution of data $p(\datsimples \vert \strs, \mathcal{D})$, the posterior distribution of NN weight $p(\latsimples_{0} \vert \strs, \mathcal{D})$, and the predictive distribution $p(\outsimples \vert \strs, \mathcal{D})$. 
$\latsimples_1$ is omitted from $\lats$ in these figures, but it behaves like $\latsimples_{0}$. 

To make a prediction, DNN uses a data point and a point-estimated NN weight. BNN uses a data point and a NN weight distribution, instead of point-estimated weight. 
VQ-BNN estimates predictive distribution by using the NN weight distribution and the set of data from the past to now that represents the probability distribution of data.
In other words, VQ-BNN without distribution of $\dats$ is equivalent to BNN, and BNN without distribution of $\lats$ is equivalent to DNN.

In this experiment, the most recent data sample is $\datsimples=0.4$. It is a noisy value because the expected value of $\datsimples$ at $t = 0$ is $0$. Since DNN and BNN only use the most recent data point to predict results, their predictive distributions are highly dependent on the noise of the data. As a result, an unexpected data makes the predictive distributions of DNN and BNN inaccurate. In contrast, VQ-BNN smoothen the predictive distribution by using predictions with respect to past data. Therefore, the predictive distributions of VQ-BNN are robust to the noise of data and its prediction.

\paragraph{Implications. }

These results imply that VQ-BNN may give a more accurate predictive result than BNN when the input and its prediction are noisy.
Also, VQ-BNN is less likely to be overconfident than BNN since VQ-BNN uses both NN weight distribution and a probability distribution of data. For this reason, VQ-BNN might be better calibrated than BNN.

\subsection{Semantic Segmentation}\label{sec:semantic-segmentation}

Semantic segmentation experiment, which is a pixel-wise classification, evaluates the computational and predictive performance of VQ-BNN with a modern deep NN in practical situation. We use the CamVid dataset \cite{brostow2009semantic} consisting of 360$\times$480 pixels 30 frame-per-second (fps) video sequences of real-world day and dusk road scenes. We use U-Net \cite{ronneberger2015u} as the backbone architecture. Bayesian U-Net, similar to \cite{kendall2015bayesian}, contains six MC dropout layers. For more information about experimental settings, see \cref{sec:setup:semantic-segmentation}. See \cref{sec:extended-semantic-segmentation} for experiments on a different dataset and model.

\paragraph{Computational performance.}

The throughput (\textsc{Bat Thr}, $\uparrow$\footnote{We use arrows to indicate which direction is better.}) column of \cref{tab:semantic-segmentation-performance} shows the number of video frames processed by each model per second in batch processing. 
In this table, VQ-BNN processes 25.5 images per second, which is only 7\% slower than DNN, and 33$\times$ faster than BNN. Likewise, the throughput column for stream processing (\textsc{Str Thr}, $\uparrow$) shows that VQ-BNN processes 9.41 images per second in stream processing, which is only 10\% slower than DNN, and 12$\times$ faster than BNN. 

In conclusion, \emph{the computational performance of VQ-BNN is comparable to that of DNN and significantly better than that of BNN.} See \cref{sec:extended-semantic-segmentation} for more information.

\paragraph{Predictive performance.}

We use global pixel accuracy (\textsc{Acc}, $\uparrow$) and mean Intersection over Union (\textsc{IoU}, $\uparrow$) to evaluate predictive results. We also use NLL ($\downarrow$), Expected Calibration Error (ECE, $\downarrow$) \cite{naeini2015obtaining, guo2017calibration}, and the following metrics to measure predictive uncertainty: 
\begin{itemize}
	\item \textbf{Accuracy-90 (\textsc{Acc$_{90}$}, $\uparrow$). } 
	If NN is confident in its prediction, it must be accurate. Therefore, we select predictions with confidence higher than 90\% and measure the accuracy, i.e., $p(\textbf{accurate} \vert \textbf{confident})$. Likewise, we measure IoU for the confident predictions (\textsc{IoU$_{90}$}, $\uparrow$).
	\item \textbf{Unconfidence-90 (\textsc{Unc$_{90}$}, $\uparrow$). } 
	If the prediction of NN is incorrect, NN should not be confident in it. 
	Therefore, we measure the probability of prediction which is not $90\%$ confident for incorrect prediction, i.e., $p(\textbf{unconfident} \vert \textbf{inaccurate})$.
	\item \textbf{Frequency-90 (\textsc{Freq$_{90}$}, $\uparrow$). } 
	Even if NN derives reliable uncertainty, the model is ineffectual if it rarely predicts high-confidence results. Therefore, we measure the percentage of predictions with $90\%$ confidence, i.e., $p(\textbf{confident})$. 
\end{itemize}

The \textsc{NLL} to \textsc{ECE} columns of the \cref{tab:semantic-segmentation-performance} show the quantitative comparison of the predictive performance for each method. 
This table shows that the predictive performance of BNN is better than that of DNN. 
Also, according to uncertainty metrics, VQ-BNN predict uncertainty better than BNN. Moreover, \textsc{Acc} and \textsc{IoU} show that VQ-BNN predicts more accurate results than BNN, which is beyond our expectations.

\begin{figure}
     \centering     
     \begin{subfigure}[b]{0.33\textwidth}
         \centering
         \includegraphics[width=\textwidth]{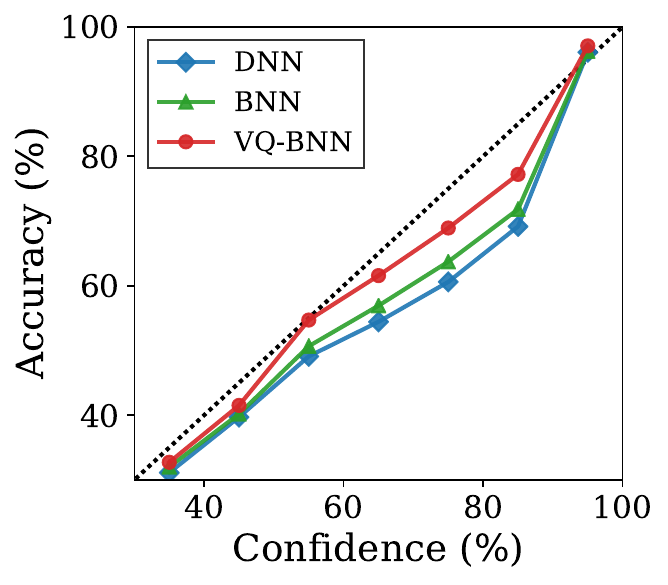}
     \end{subfigure}
     
\vskip -0.05in
     \caption{\textbf{Reliability diagram} with semantic segmentation. The black dotted line shows the accuracy we expect for each confidence.}
\vskip -0.13in
     \label{fig:reliability-diagram}
\end{figure}

\Cref{fig:reliability-diagram} shows the reliability diagram \cite{niculescu2005predicting, naeini2015obtaining, guo2017calibration}. As shown in this figure, DNN is miscalibrated; there is significant discrepancy between confidence and accuracy. In contrast, VQ-BNN is better calibrated than DNN, and surprisingly better than BNN.

According to these results, \emph{VQ-BNN is the most appropriate method not only to distinguish uncertain predictions but also to predict accurate results}. \Cref{tab:extended-semantic-segmentation} in \cref{sec:extended-semantic-segmentation} shows the predictive performance of VQ-BNN in various situations.
\cref{sec:extended-semantic-segmentation} also evaluates VQ-DNN, which is the temporal smoothing of DNN's predictions. Its predictive performance is better than that of DNN, but worse than that of VQ-BNN.

\paragraph{Analysis. }

The results of \cref{sec:simple-linear-regression} imply that VQ-BNN is effective in compensating for noisy predictions. For semantic segmentation task, the data that derives inaccurate results mainly correspond to the edges of objects. 

\begin{figure*}
     \centering
     
     \begin{subfigure}[b]{\textwidth}
         \centering
         \includegraphics[width=0.98\textwidth,page=1]{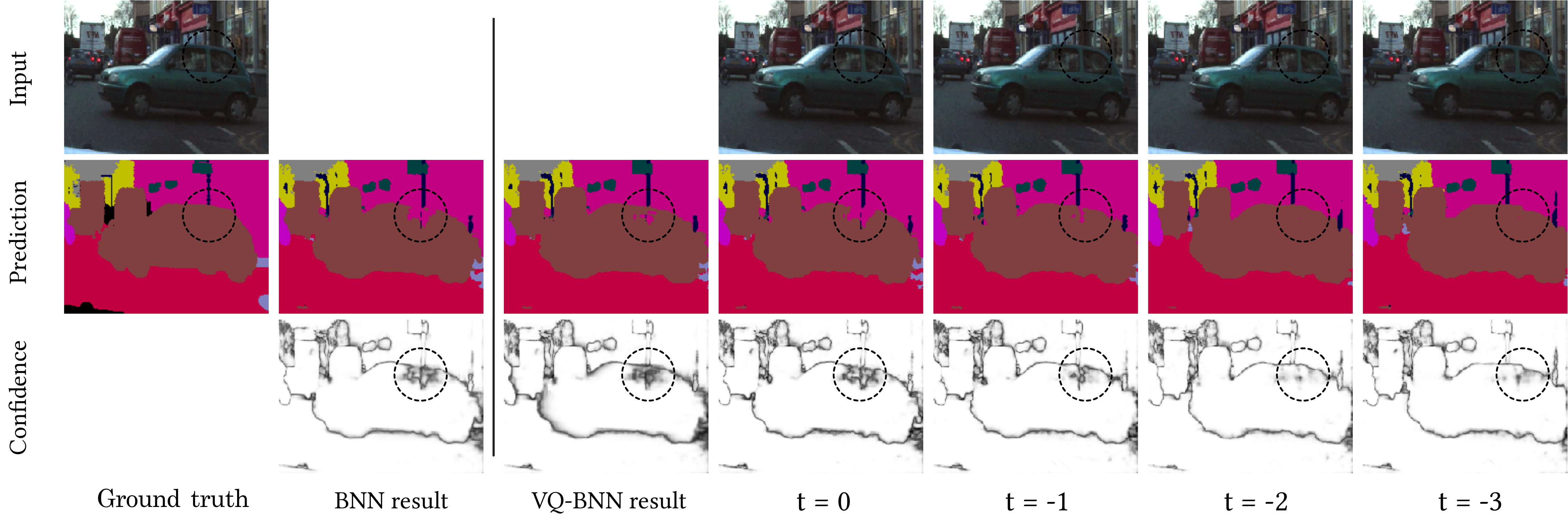}
     \end{subfigure}

	\vskip -0.1in
     \caption{
     \textbf{Qualitative Analysis of VQ-BNN} with semantic segmentation. 
     The first row is (cropped and adjusted) input image, the second row is prediction for the image, and the last row is confidence. A whiter background corresponds to higher confidence.
     VQ-BNN predicts the result once for the most recent image ($t=0$). Then, it derives the predictive distribution by adding the latest prediction ($t=0$) and past predictions ($t\in\{ -1, -2, \cdots \}$), with exponentially decaying importances.
     Since the objects in the video sequence move slowly, the predictive distribution of VQ-BNN has high uncertainty at the edges of the objects.
     Even if VQ-BNN accidentally predicts a wrong result for the most recent frame, the past predictions compensate for this error. 
     In this figure, the predictive uncertainty of VQ-BNN is mainly located at the edge of the car, and VQ-BNN derives more accurate predictive result for the car than BNN does.
     }
     \label{fig:qualitative-analysis}
\end{figure*}

VQ-BNN smoothens the predictive distribution by using past predictions. Because objects move slowly every frame, the uncertainty predicted by VQ-BNN is located at the edges of the objects. Even if VQ-BNN accidentally predicts a wrong result for the most recent frame, the past predictions compensate for this error. \Cref{fig:qualitative-analysis} shows that predictive uncertainty of VQ-BNN is mainly located at the edge of the car, and past predictions correct the most recent incorrect predictions for the car.

We quantitatively show that VQ-BNN achieves higher predictive performance than other methods at the edges of objects. 
We measure predictive performance for pixels representing the object edges, and we call it \emph{edge predictive performance}. 
The results of this experiment show that the edge predictive performance lags behind the predictive performance for all pixels. It implies that there are many inaccurate predictions on object edges.
In addition, the difference in the edge predictive performance between VQ-BNN and BNN is greater than the difference in the predictive performance between VQ-BNN and BNN. This implies that VQ-BNN works well for edge pixels. 
See \cref{sec:extended-semantic-segmentation} and \cref{tab:edge-performance} for more details on the edge predictive performance.

VQ-BNN relies on temporal consistency of data streams. \cref{sec:extended-semantic-segmentation} evaluate the sensitivity to the temporal consistency, and the result show that the predictive performance is degraded when temporal consistency decreases.

\subsection{Depth Estimation}\label{sec:depth-estimation}

\begin{table}[t]
\vskip 0.1in
\begin{center}
\begin{small}
\begin{sc}

  \begin{tabular}{lcccc}
    \toprule
    Method & \thead{Bat Thr\\(Img/Sec)} & \thead{Str Thr\\(Img/Sec)} & NLL & \thead{RMSE\\(m)} \\
    \midrule
    DNN    & \textbf{54.0} & \textbf{14.5} & 1.55 & 0.804 \\
    BNN    & 1.59 & 1.61 & 1.10 & 0.705 \\
    VQ-BNN & 50.8 & 13.6 & \textbf{1.09} & \textbf{0.700} \\
    
    \bottomrule
  \end{tabular}

\end{sc}
\end{small}
\end{center}
\vskip -0.05in
\caption{
\textbf{Computational and predictive performance with depth estimation} for each method.
}
\label{tab:depth-estimation-performance}
\vskip -0.1in
\end{table}

Monocular depth estimation experiment shows the performance of VQ-BNN with a deep NN in a regression task on a real-world dataset. We use the NYUDv2 dataset \cite{Silberman:ECCV12}, which consists of 240$\times$320 pixels 20-30 fps video sequences from a variety of indoor scenes. As in \cref{sec:semantic-segmentation}, we use U-Net and Bayesian U-Net as backbone architectures. For more information about experimental settings, see \cref{sec:setup:depth-estimation}.

\paragraph{Computational performance. }

The throughput for batch processing (\textsc{Bat Thr}, $\uparrow$) of \cref{tab:depth-estimation-performance} shows the number of video frames processed by each model per second. 
In this table, VQ-BNN processes 54.0 images per second, which is only 6\% slower than DNN, and 32$\times$ faster than BNN.
Similarly, the throughput for stream processing (\textsc{Str Thr}, $\uparrow$) shows that VQ-BNN processes 13.6 images per second in stream processing, which is only 6\% slower than that of DNN, and 8$\times$ faster than that of BNN. 
These results are consistent with the results in \cref{sec:semantic-segmentation}; the computational performance of VQ-BNN is significantly better than that of BNN, and is similar to that of DNN. See \cref{sec:extended-depth-estimation} for more information.

\paragraph{Predictive performance. }

We use root-mean-square error (\textsc{RMSE}, $\downarrow$) to evaluate predictive results for depth estimation. We use \textsc{NLL} to evaluate predictive uncertainty. 

\textsc{RMSE} and \textsc{NLL} columns of \cref{tab:depth-estimation-performance} show predictive performances for each method. This table shows that both RMSE and NLL of VQ-BNN are the lowest among those of the three methods.
In conclusion, VQ-BNN is the most appropriate method for predicting accurate predictive results as well as reliable uncertainty in the regression task. See \cref{sec:extended-depth-estimation} for more information on depth regression experiment.

\section{Related Work}\label{sec:related-work}

Several sampling-free BNNs, e.g. \cite{HernandezLobato:2015vo, wang2016natural, wu2018deterministic}, were proposed recently and they might be a solution to the problem that BNNs require multiple NN predictions. Sampling-free BNNs approximate the posterior and the probability of layer's outputs using a simple type of parametric distribution such as Gaussian distribution or exponential family. Therefore, they predict results with one or two forward passes. 

However, neural networks used in real world situations have dozens or more of layers, and sampling-free BNNs are not suitable for deep NNs. To the best of our knowledge, most sampling-free BNN have been evaluated only with a couple of layers on low-dimensional data such as UCI datasets.
One of the reasons is that the Gaussian approximation in sampling-free BNNs can be inaccurate to represent real-world probabilities. Since the discrepancy between true values and approximate values accumulates in every NN layer, the error of deep sampling-free BNNs becomes not negligible.
Moreover, in many cases, sampling-free BNN can not use variational inference to obtain a posterior because ELBO is not amenable. 
For these reasons, we mainly consider sampling-based BNNs for comparison in this paper. 
Recently, \cite{gast2018lightweight} and \cite{haussmann2020sampling} applied sampling-free BNNs to LeNet. \cite{postels2019sampling} applied it to SegNet; however, the neural network does not predict well-calibrated results. See \cref{sec:sampling-free} for more details on the sampling-free BNN.

\cite{riquelme2018deep} and \cite{kohl2018probabilistic}, which utilize Bayesian methods only in the last layer, predict results efficiently. However, this approach generally achieves poor predictive performance and are not robust to corrupted inputs \cite{ovadia2019can}.

A temporal smoothing has been widely used to reduce noise for accurate time-series forecasting \cite{pai2005hybrid,ediger2007arima,benvenuto2020application}.  \cite{zhang2003time,khashei2010artificial,chan2011neural} combined it with NN to improve accuracy for forecasting tasks on low-dimensional data streams. In this paper, we show that the temporal smoothing can significantly improve the computational performance of BNNs on high-dimensional data streams.

\section{Conclusion}

We present VQ-BNN inference, which is a novel approximation of BNN inference, to improve the computational performance of BNN inference for data streams. BNN inference iteratively executes NN prediction for a data, which makes it dozens of times slower. In contrast, VQ-BNN inference executes NN prediction only once for the latest data from the data stream, and compensate the result with previously memorized predictions. Specifically, VQ-BNN inference for data streams is temporal smoothing of recent predictions with exponentially decaying importance, and it is easy to implement. This method results in an order of magnitude times improvement in computational performance compared to BNN. 
Experiments with computer vision tasks such as semantic segmentation on various real-world datasets show that the computational performance of VQ-BNN is almost the same as that of deterministic NN, and the predictive performance is comparable to or even superior to that of BNN. 
Since the computational performance of deterministic NN is the best we can expect, VQ-BNN is an efficient method to estimate uncertainty.
 
\section*{Acknowledgement}

This work was supported by Samsung Research Funding Center of Samsung Electronics under Project Number SRFC-IT1801-10.

\bibliography{aaai}

\clearpage

\appendix

\section{Experimental Setup and Datasets}\label{sec:setup}

\begin{table}[t]
\vskip 0.1in
\begin{center}
\small
\begin{sc}

  \begin{tabular}{lccc}
    \toprule
    {Dataset} 		& {$\dim{(\dats)}$} & {$\dim{(\outs)}$} & {$\vert \mathcal{D} \vert$} \\
    \midrule
    CamVid			& 360$\times$480$\times$3 	& 360$\times$480$\times$11 & 421  \\
    CityScape		& 512$\times$1024$\times$3 	& 512$\times$1024$\times$19 & 2975  \\
    \midrule
    
    NYUDv2			& 240$\times$320$\times$3 	& 240$\times$320$\times$1 & 47584  \\
    \bottomrule
  \end{tabular}

\end{sc}
\end{center}
\caption{
\textbf{Informations about the datasets.} 
It represents input dimensionality ($\dim{(\dats)}$), output dimensionality ($\dim{(\outs)}$), and cardinality of training set ($\vert \mathcal{D} \vert$) of dataset used in the experiments.
}
\label{tab:datasets}

\vskip -0.1in
\end{table}

We conduct all the experiments with the Intel Xeon W-2123 Processor, 32GB memory, and a single GeForce RTX 2080 Ti. NN models are implemented in TensorFlow \cite{abadi2016tensorflow}.

\subsection{Semantic Segmentation}\label{sec:setup:semantic-segmentation}

\cref{sec:semantic-segmentation}, \cref{sec:extended-semantic-segmentation}, and \cref{sec:sampling-free} conduct semantic segmentation experiments. In these experiments, we use two different kinds of neural network architectures: U-Net \cite{ronneberger2015u} based on VGG-16 \cite{simonyan2014very} and SegNet \cite{badrinarayanan2017segnet}. 
We also use Bayesian U-Net and Bayesian SegNet \cite{kendall2015bayesian} as baselines. Bayesian U-Net, similar to \cite{kendall2015bayesian}, contains six MC dropout \cite{Gal:2016uw} layers behind the layers that receive the smallest feature map sizes. 

We use the CamVid dataset \cite{brostow2009semantic} and the CityScape dataset \cite{cordts2016cityscapes} in the semantic segmentation experiments. For the CamVid dataset, we resize images to 360$\times$480 pixels bilinearly from 720$\times$960 pixels. For the CityScape dataset, we resize images to 512$\times$1024 pixels from 1024$\times$2048 pixels. For the CityScape dataset, we use validation set as test set because only the video sequence corresponding to the validation set has been disclosed. We use reduced label lists on both datasets. \cref{tab:datasets} summarizes the information of datasets. 

NNs are trained using categorical cross-entropy loss, with Adam optimizer with a constant learning rate of 0.001, $\beta_{1} = 0.9$, and  $\beta_{2} = 0.999$. Batch size is limited to 3 because of memory limitation. We consider basic data augmentation techniques: random cropping and horizontal flipping. Since the datasets are imbalanced, we consider median frequency balancing. For the CamVid dataset and the CityScape dataset, we train NNs for 100 epochs and 500 epochs, respectively.

We set $p(\outs \vert \inps_{i})$ as a categorical distribution parameterized by the softmax of NN logit, then the predictive distribution of BNN is $p(\outs \vert \dats_{0}, \mathcal{D}) \simeq \sum_{i} \frac{1}{N} \, \texttt{Softmax}(\nn (\dats_{0}, \lats_{i}))$ where $N$ is number of NN weight samples and $\nn (\cdot)$ is logit of NN. The predictive distribution of VQ-BNN is \cref{eq:vqbnn-classification}.
The reported metrics are the mean values and standard deviations of five evaluations. For the CityScape, the metrics are the mean and standard deviations from the final five epochs. To optimize the predictive performances, we set hyperparameters of VQ-BNN $K$ and $\tau$ to $5$ and $1.25$, respectively. We report how NLL changes across various hyperparameters in \cref{app:hyperparameters}.

\begin{figure}
     \centering     
     \begin{subfigure}[b]{0.30\textwidth}
         \centering
         \includegraphics[width=\textwidth]{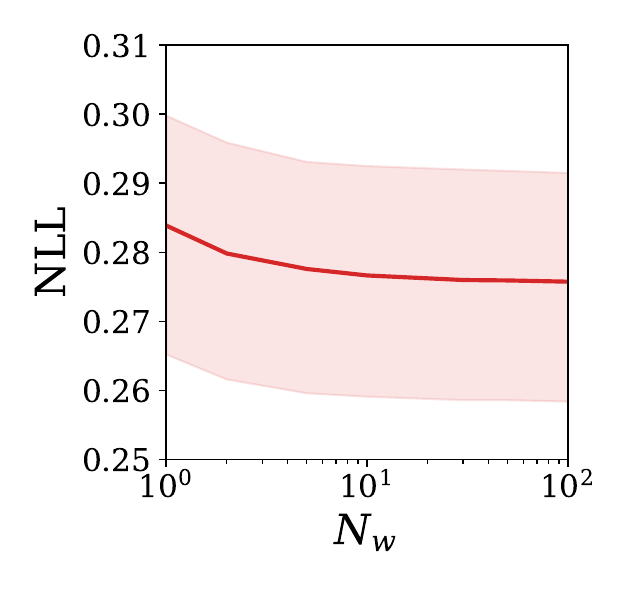}
         \label{fig:confidence:dnn}
     \end{subfigure}
	\vskip -0.3in
	
     \caption{
     \textbf{NLL of BNN with semantic segmentation for the number of forward passes} ($N_{\lats}$).
     }
     \label{fig:performance-sample-number}
\end{figure}

\subsection{Depth Estimation}\label{sec:setup:depth-estimation}

\cref{sec:depth-estimation} and \cref{sec:extended-depth-estimation} conduct monocular depth estimation experiments. In these experiment, we use the same U-Net, SegNet, Bayesian U-Net, and Bayesian SegNet as above. 

We use NYU Depth Dataset V2 (NYUDv2) \cite{Silberman:ECCV12} that resizes 480$\times$640 pixels to 240$\times$320 pixels in the experiments. NYUDv2 contains 464 scenes, and 249 scenes are used for training. We use the remaining 215 scenes for evaluation.

NNs are optimized with mean squared error loss function and Adam optimizer with a constant learning rate of 0.001, $\beta_{1} = 0.9$, and  $\beta_{2} = 0.999$. Batch size is limited to 3. We consider simple data augmentation techniques: random cropping and horizontal flipping. We train NN for 6 epochs.

Since depth estimation is a pixel-wise regression task, we set $p(\outs \vert \inps_{i})$ as a normal distribution $\mathcal{N}(\outs \vert \nn (\dats_{0}, \lats_{i}), \sigma^{2})$ where $\sigma$ is hyperparameter. Then, the predictive distribution of BNN is $p(\outs \vert \dats_{0}, \mathcal{D}) \simeq \sum_{i} \frac{1}{N} \mathcal{N}(\outs \vert \nn (\dats_{0}, \lats_{i}), \sigma^{2})$. The predictive distribution of VQ-BNN is given by \cref{eq:vqbnn-regression} in theory, but this expression is a bit tricky to implement. Instead, we rewrite the distribution as $\outs' + \bm{\epsilon}$ where $\outs' \sim \sum_{t=0}^{-K} \pi(\dats_{t} \vert \strs) \, \delta(\outs - \nn (\dats_{t}, \lats_{t}))$ and $\bm{\epsilon} \sim \mathcal{N}(0, \sigma^{2})$. Then, we use normal distribution to approximate the distribution of $\outs'$ as follows: 
\begin{align}
	p(\outs \vert \strs, \mathcal{D}) 
	&\simeq \sum_{t=0}^{-K} \pi(\dats_{t} \vert \strs) \, \mathcal{N}(\outs \vert \nn (\dats_{t}, \lats_{t}), \sigma^{2}) \\
	&\simeq \mathcal{N}(\mu', {\sigma'}^{2} + {\sigma}^{2})
	\label{eq:vqbnn-regression-simplified}
\end{align}
where $\mu'$ and $\sigma'^{2}$ are weighted mean and weighted variance of $\{ \nn (\dats_{t}, \lats_{t}) \}$, i.e.,
\begin{align}
	\mu^{\prime} &= \sum_{t=0}^{-K} \pi(\dats_{t} \vert \strs) \nn (\dats_{t}, \lats_{t})\\
	{\sigma^{\prime}}^{2} &= \sum_{t=0}^{-K} \pi(\dats_{t} \vert \strs) \left( \nn (\dats_{t}, \lats_{t}) - \mu^{\prime} \right)^{2}
\end{align}
The reported metrics are the mean and standard deviations from $5 + \sfrac{1}{3}$, $5 + \sfrac{2}{3}$, and $6$ epochs. We set $\sigma$ to $0.4$, which is the value that minimizes NLL on validation set. We set hyperparameters of VQ-BNN $K$ and $\tau$ to $5$ and $1.25$, respectively, as above.

\section{Predictive Performance of BNN for Different Numbers of Forward Passes}\label{sec:bnn-performance-for-forward-pass}

BNNs in experiments contain MC dropout layers and require multiple forward passes to predict result. \Cref{fig:performance-sample-number} shows the predictive performance represented by NLL of the Bayesian U-Net for the number of forward passes on the CamVid dataset.

This figure shows that the predictive performance of BNN improves up to 30 forward passes. However, there is little difference between the predictive performance for 30 forward passes and 50 forward passes. According to this result, BNN requires at least 30 forward passes to get the best predictive performance, so we execute 30 predictive inference in all experiments. 

\section{Effects of Hyperparameters} \label{app:hyperparameters}

\begin{figure}
     \centering
     \begin{subfigure}[b]{0.23\textwidth}
         \centering
         \includegraphics[width=\textwidth]{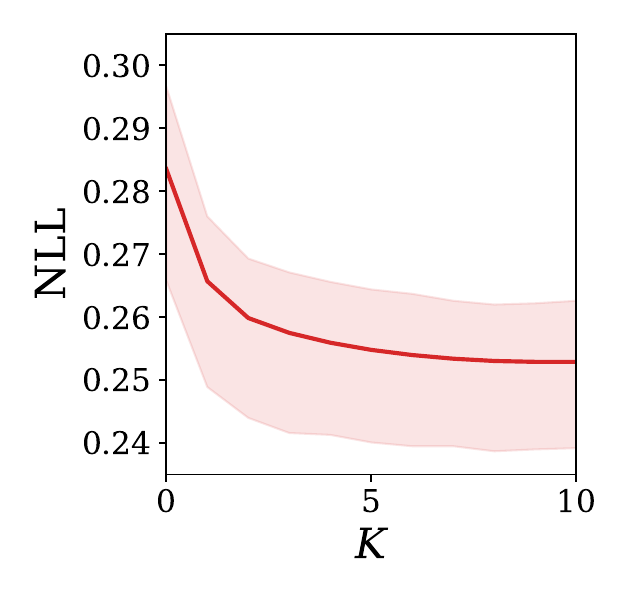}
     \end{subfigure}
     \begin{subfigure}[b]{0.23\textwidth}
         \centering
         \includegraphics[width=\textwidth]{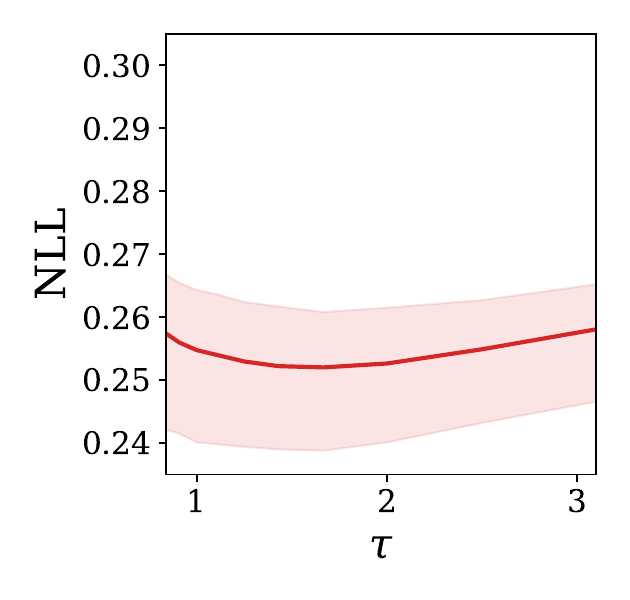}
     \end{subfigure}
	\vskip -0.1in
     
     \caption{
     \textbf{NLLs of VQ-BNN for hyperparameters $K$ and $\tau$.}
     }
     \label{fig:hyperparameters}
\end{figure}

VQ-BNN has two independent hyperparameters: $K$ and $\tau$. \Cref{fig:hyperparameters} shows how NLL changes over the variuos hyperparameters in the semantic segmentation experiment on the CamVid dataset. As shown in this figure, as $K$ increases, NLL decreases, and the NLL stagnates when $K$ goes over 5. The higher the $\tau$, the lower the NLL, until 1.25.

With high $K$ and $\tau$, VQ-BNN uses more previous predictions to derive the predictive distribution. In this case, past predictions compensate for the most recent prediction as shown in this paper. As a result, with high $K$ and $\tau$, VQ-BNN exhibits lower NLL. 
However, when $\tau$ gets higher than a certain point ($\tau = 1.25$ in this figure), the data uncertainty is too high to estimate results. NLL is minimized when model uncertainty and data uncertainty are balanced.

\section{Extended Informations of Experiments}\label{sec:extended-experiments}

\cref{sec:experiment} makes a computer vision experiments namely semantic segmentation and depth estimation to show the performance of VQ-BNN. This section shows additional information on these experiments.

We use the additional method in the experiments:
\begin{itemize}
	\item \textbf{VQ-DNN.} 
	VQ-DNN is temporal exponential smoothing of DNN's predictions. In other words, it is almost the same as VQ-BNN inference, but it uses deterministic NN weight $p(\lats \vert \mathcal{D}) = \delta(\lats - \lats_{0})$ obtained from DNN, instead of probabilistic NN weights. This method shows the effect of data uncertainty, when it is applied to the DNN. This technique is particularly useful when BNN is not accessible. The experimental results show that VQ-DNN achieves better predictive performance than vanilla DNNs in computer vision tasks.
\end{itemize}

\subsection{Semantic Segmentation}\label{sec:extended-semantic-segmentation}

\begin{figure}
     \centering     
     \begin{subfigure}[b]{0.33\textwidth}
         \centering
         \includegraphics[width=\textwidth]{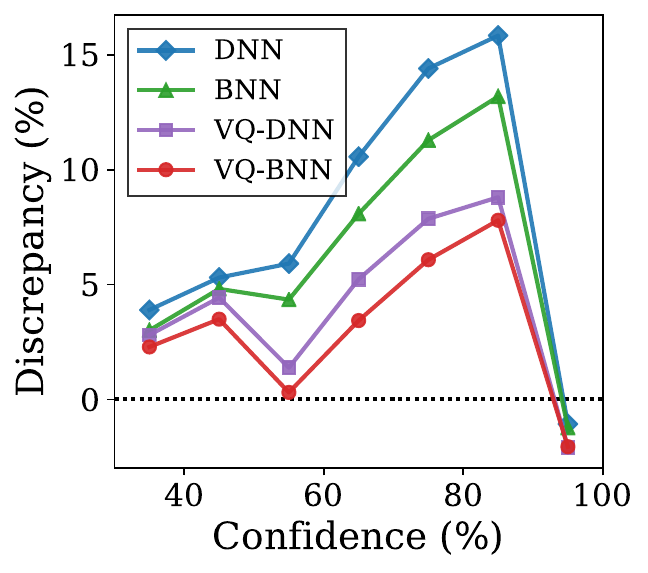}
     \end{subfigure}
     
     \caption{\textbf{Reliability diagram} with semantic segmentation on the CamVid dataset. Discrepancy is defined as confidence minus accuracy. The black dotted lines show the discrepancy we expect.}
     \label{fig:reliability-diagram-extended}
\end{figure}

\begin{sidewaystable*}

\small
\centering
\begin{sc}
  \begin{tabular}{llcccccccccc}
    \toprule
    
    \thead{Dataset \& \\Model} & Method & \thead{Bat Thr\\(Img/Sec)} & \thead{Str Thr\\(Img/Sec)} & NLL & \thead{Acc\\(\%)} & \thead{Acc$_{90}$\\(\%)} & \thead{Unc$_{90}$\\(\%)} & \thead{IoU\\(\%)} & \thead{IoU$_{90}$\\(\%)} & \thead{Freq$_{90}$\\(\%)} & \thead{ECE\\(\%)} \\
    \midrule

    \multirow{4}{*}{\thead{CamVid \&\\U-Net}}
    & DNN & \textbf{27.5$\pm$1.2} & \textbf{10.5$\pm$1.7}
    & 0.314$\pm$0.021 & 91.1$\pm$1.3 & 96.1$\pm$0.3 & 61.3$\pm$6.4 & 66.1$\pm$2.7 & 77.7$\pm$0.6 & 86.4$\pm$3.5 & 4.31$\pm$0.37 \\
    & BNN & 0.824$\pm$0.003 & 0.788$\pm$0.033 
    & 0.276$\pm$0.012 & 91.8$\pm$0.3 & 96.5$\pm$0.2 & 63.0$\pm$3.0 & 68.1$\pm$0.7 & 79.9$\pm$0.5 & \textbf{86.8$\pm$1.2} & 3.71$\pm$0.23 \\
    & VQ-DNN & 27.0$\pm$1.6 & 10.3$\pm$1.3 
    & 0.284$\pm$0.023 & 91.2$\pm$1.2 & 97.0$\pm$0.2 & 71.0$\pm$5.2 & 66.6$\pm$2.7 & 81.7$\pm$0.6 & 83.0$\pm$3.5 & 3.00$\pm$0.33 \\
    & VQ-BNN & 25.5$\pm$1.1 & 9.41$\pm$0.84 
    & \textbf{0.253$\pm$0.009} & \textbf{92.0$\pm$0.2} & \textbf{97.4$\pm$0.2} & \textbf{72.5$\pm$2.8} & \textbf{68.6$\pm$0.8} & \textbf{83.7$\pm$0.8} & 83.1$\pm$1.3 & \textbf{2.24$\pm$0.30} \\
    \midrule

    \multirow{4}{*}{\thead{CamVid \&\\SegNet}}
    & DNN & \textbf{37.0$\pm$1.5} & \textbf{11.9$\pm$1.6}
    & 0.605$\pm$0.090 & 86.2$\pm$0.7 & 92.6$\pm$1.2 & 55.1$\pm$6.4 & 51.6$\pm$1.3 & 60.9$\pm$3.1 & \textbf{83.7$\pm$2.3} & 8.38$\pm$1.17 \\
    & BNN & 1.13$\pm$0.01 & 1.12$\pm$0.03
    & \textbf{0.426$\pm$0.026} & \textbf{86.8$\pm$0.8} & 96.1$\pm$0.5 & 78.5$\pm$2.1 & \textbf{53.7$\pm$1.1} & 71.2$\pm$1.8 & 73.6$\pm$1.5 & 4.02$\pm$0.75 \\
    & VQ-DNN & 35.8$\pm$2.6 & 11.4$\pm$1.3 
    & 0.483$\pm$0.099 & 86.7$\pm$0.8 & 94.6$\pm$1.3 & 68.3$\pm$6.6 & 52.3$\pm$1.4 & 65.3$\pm$4.1 & 78.8$\pm$2.5 & 6.14$\pm$1.38 \\
    & VQ-BNN & 34.8$\pm$2.0 & 11.3$\pm$1.2 
    & 0.433$\pm$0.027 & 86.2$\pm$0.9 & \textbf{96.5$\pm$0.9} & \textbf{81.8$\pm$5.6} & 52.7$\pm$1.2 & \textbf{72.4$\pm$2.2} & 70.6$\pm$5.9 & \textbf{3.55$\pm$1.71} \\
    \midrule

    \multirow{4}{*}{\thead{CityScape \&\\U-Net}}
    & DNN & \textbf{9.20$\pm$0.22} & \textbf{7.44$\pm$0.80}
    & 0.325$\pm$0.057 & 91.4$\pm$1.6 & 96.2$\pm$0.4 & 60.9$\pm$4.9 & 60.0$\pm$5.9 & 71.7$\pm$4.1 & \textbf{87.5$\pm$3.9} & 4.30$\pm$0.29 \\
    & BNN & 0.285$\pm$0.003 & 0.285$\pm$0.003 
    & \textbf{0.278$\pm$0.051} & \textbf{92.4$\pm$1.3} & 97.3$\pm$0.3 & 68.5$\pm$4.0 & \textbf{60.7$\pm$3.9} & 75.3$\pm$1.7 & 86.8$\pm$3.8 & 2.95$\pm$0.23 \\
    & VQ-DNN & 9.09$\pm$0.34 & 7.39$\pm$0.68 
    & 0.296$\pm$0.050 & 91.7$\pm$1.5 & 97.5$\pm$0.4 & 75.5$\pm$5.2 & 60.6$\pm$5.7 & 75.5$\pm$3.9 & 78.7$\pm$5.8 & 2.00$\pm$0.56 \\
    & VQ-BNN & 8.59$\pm$0.25 & 6.53$\pm$0.50 
    & 0.282$\pm$0.048 & 92.2$\pm$1.5 & \textbf{97.9$\pm$0.2} & \textbf{78.1$\pm$3.1} & 60.4$\pm$3.9 & \textbf{76.5$\pm$1.5} & 80.0$\pm$3.9 & \textbf{1.53$\pm$0.23} \\
    \midrule

    \multirow{4}{*}{\thead{CityScape \&\\SegNet}}
    & DNN & \textbf{12.4$\pm$0.7} & \textbf{8.89$\pm$0.55}
    & 0.361$\pm$0.011 & 89.6$\pm$0.4 & 96.0$\pm$0.2 & 67.6$\pm$1.0 & 54.4$\pm$0.6 & 69.3$\pm$0.5 & \textbf{83.9$\pm$0.5} & 4.80$\pm$0.29 \\
    & BNN & 0.393$\pm$0.006 & 0.393$\pm$0.005
    & 0.328$\pm$0.004 & 89.4$\pm$0.2 & 97.8$\pm$0.1 & 84.4$\pm$0.3 & 53.3$\pm$0.3 & \textbf{77.6$\pm$0.4} & 75.8$\pm$0.3 & 1.79$\pm$0.10 \\
    & VQ-DNN & 12.3$\pm$0.8 & 8.77$\pm$0.49 
    & \textbf{0.327$\pm$0.008} & \textbf{90.4$\pm$0.3} & 97.6$\pm$0.1 & 80.8$\pm$0.6 & \textbf{55.0$\pm$0.5} & 73.5$\pm$0.3 & 77.3$\pm$0.7 & 1.88$\pm$0.11 \\
    & VQ-BNN & 11.7$\pm$0.6 & 8.05$\pm$0.677
    & 0.344$\pm$0.004 & 89.1$\pm$0.2 & \textbf{98.1$\pm$0.1} & \textbf{87.7$\pm$0.2} & 52.4$\pm$0.2 & 77.3$\pm$0.6 & 71.3$\pm$0.3 & \textbf{0.784$\pm$0.040} \\
    
    \bottomrule
  \end{tabular}
\end{sc}

\caption{
\textbf{Computational and predictive performance with semantic segmentation} for each method.
}
\label{tab:extended-semantic-segmentation}
\vskip 0.3in

\small
\centering
\begin{sc}
  \begin{tabular}{llccccccccc}
    \toprule
    
    \thead{Dataset \& \\Model} & Method & \thead{Bat Thr\\(Img/Sec)} & \thead{Str Thr\\(Img/Sec)} & NLL & \thead{RMSE\\(m)} & \thead{Rel} & \thead{$\delta < 1.25^{1}$} & \thead{$\delta < 1.25^{2}$} & \thead{$\delta < 1.25^{3}$} \\
    \midrule

    \multirow{4}{*}{\thead{NYUDv2 \&\\U-Net}}
    & DNN & \textbf{54.0$\pm$2.6} & \textbf{14.5$\pm$2.1}
    & 1.55$\pm$0.36 & 0.804$\pm$0.073 & 0.298$\pm$0.034 & 0.560$\pm$0.029 & 0.838$\pm$0.011 & 0.939$\pm$0.010 \\
    & BNN & 1.59$\pm$0.02 & 1.61$\pm$0.03 
    & 1.10$\pm$0.03 & 0.704$\pm$0.018 & 0.293$\pm$0.012 & 0.573$\pm$0.023 & 0.834$\pm$0.016 & 0.941$\pm$0.005 \\
    & VQ-DNN & 52.8$\pm$2.8 & 13.9$\pm$1.5
    & 1.14$\pm$0.07 & 0.750$\pm$0.081 & 0.299$\pm$0.037 & 0.565$\pm$0.034 & 0.838$\pm$0.016 & 0.940$\pm$0.012 \\
    & VQ-BNN & 50.8$\pm$3.1 & 13.6$\pm$1.4
    & \textbf{1.09$\pm$0.03} & \textbf{0.700$\pm$0.016} & \textbf{0.289$\pm$0.007} & \textbf{0.581$\pm$0.014} & \textbf{0.839$\pm$0.010} & \textbf{0.942$\pm$0.003} \\
    \midrule

    \multirow{4}{*}{\thead{NYUDv2 \&\\SegNet}}
    & DNN & \textbf{68.8$\pm$0.3} & \textbf{15.0$\pm$0.3}
    & 1.18$\pm$0.06 & 0.740$\pm$0.028 & 0.335$\pm$0.026 & 0.529$\pm$0.031 & 0.807$\pm$0.024 & 0.925$\pm$0.009 \\
    & BNN & 2.07$\pm$0.00 & 2.09$\pm$0.00 
    & 1.19$\pm$0.09 & 0.762$\pm$0.045 & 0.348$\pm$0.024 & 0.518$\pm$0.013 & 0.797$\pm$0.017 & 0.917$\pm$0.011 \\
    & VQ-DNN & 67.6$\pm$0.7 & 14.9$\pm$0.2
    & \textbf{1.13$\pm$0.08} & \textbf{0.717$\pm$0.042} & \textbf{0.325$\pm$0.032} & \textbf{0.538$\pm$0.033} & \textbf{0.813$\pm$0.027} & \textbf{0.929$\pm$0.010} \\
    & VQ-BNN & 69.6$\pm$0.1 & 14.2$\pm$0.2
    & 1.15$\pm$0.01 & 0.744$\pm$0.014 & 0.329$\pm$0.020 & 0.519$\pm$0.006 & 0.812$\pm$0.015 & 0.927$\pm$0.011 \\

    \bottomrule
  \end{tabular}
\end{sc}

\caption{
\textbf{Computational and predictive performance with depth estimation} for each method.
}
\label{tab:extended-depth-estimation}

\end{sidewaystable*}

\Cref{sec:semantic-segmentation} shows the performance of U-Net with semantic segmentation on the CamVid dataset. This section uses SegNet additionally, and shows the performances on the CityScape dataset, which is another real-world road scene video sequence. Furthermore, we analyze the predictive performances of DNN, BNN, VQ-DNN, and VQ-BNN in more details.

\begin{table*}[t]
\vskip 0.1in
\begin{center}
\begin{small}
\begin{sc}
  \begin{tabular}{clccccccccc}
    \toprule
    
    \thead{Threshold\\(Percentage, \%)} & Method & NLL & \thead{Acc\\(\%)} & \thead{Acc$_{90}$\\(\%)} & \thead{Unc$_{90}$\\(\%)} & \thead{IoU\\(\%)} & \thead{IoU$_{90}$\\(\%)} & \thead{Freq$_{90}$\\(\%)} & \thead{ECE\\(\%)} \\
    \midrule
    
    \multirow{4}{*}{\thead{0.0\\(100\%)}}
    & DNN & 0.314 & 91.1 & 96.1 & 61.3 & 66.1 & 77.7 & 86.4 & 4.31  \\
    & BNN & 0.276 & 91.8 & 96.5 & 63.0 & 68.1 & 79.9 & \textbf{86.8} & 3.71  \\
    & VQ-DNN & 0.284 & 91.3 & 97.1 & 72.1 & 66.6 & 82.1 & 82.4 & 2.72  \\
    & VQ-BNN & \textbf{0.256} & \textbf{92.0} & \textbf{97.3} & \textbf{72.3} & \textbf{68.5} & \textbf{83.5} & 83.0 & \textbf{2.27}  \\
    \midrule

    \multirow{4}{*}{\thead{0.5\\(14.2\%)}}
    & DNN & 0.599 & 82.6 & 91.5 & 64.0 & 55.3 & 68.6 & \textbf{73.0} & 8.47  \\
    & BNN & 0.530 & 83.7 & 92.7 & 67.4 & 56.4 & 70.7 & 72.5 & 6.53  \\
    & VQ-DNN & 0.546 & 82.9 & 93.5 & 74.5 & 55.7 & 73.6 & 66.1 & 5.56  \\
    & VQ-BNN & \textbf{0.492} & \textbf{84.1} & \textbf{94.3} & \textbf{76.3} & \textbf{57.1} & \textbf{75.7} & 66.6 & \textbf{4.64}  \\
    \midrule

    \multirow{4}{*}{\thead{1.0\\(5.76\%)}}
    & DNN & 0.665 & 80.9 & 90.1 & 62.7 & 52.7 & 64.3 & \textbf{71.6} & 9.66  \\
    & BNN & 0.588 & 82.0 & 91.5 & 66.7 & 53.6 & 66.6 & 70.8 & 8.28  \\
    & VQ-DNN & 0.604 & 81.3 & 92.4 & 74.1 & 53.3 & 68.4 & 63.4 & 6.21  \\
    & VQ-BNN & \textbf{0.543} & \textbf{82.4} & \textbf{93.6} & \textbf{77.0} & \textbf{54.2} & \textbf{70.2} & 62.8 & \textbf{4.93}  \\

    \bottomrule
  \end{tabular}
\end{sc}
\end{small}
\end{center}

\caption{
\textbf{Predictive performance with semantic segmentation for edge pixels.}
}
\label{tab:edge-performance}
\vskip -0.1in
\end{table*}

\paragraph{Computational and predictive performances. }

\Cref{tab:extended-semantic-segmentation} shows the computational and predictive performance of various methods with U-Net and SegNet on the CamVid and the CityScape dataset. As in \cref{sec:semantic-segmentation}, we measure the metrics for predictive results and predictive uncertainties.

In general, we improves the computational performance with larger batch size. In batch processing, we use batch sizes of 10 and 5 on the CamVid and the CityScape dataset, respectively. 
In contrast, in stream processing, data is given one by one; we predict results for one input data per execution. DNN, VQ-DNN, and VQ-BNN use a batch size of 1 to predict a result for one data. Therefore, the throughput of these methods in stream processing is lower than that in batch processing. BNN uses a batch size of 10 or 5 since it executes 30 NN predictions to derive a result for one data. Thus, the throughput of BNN is almost the same in both situations.
Nevertheless, BNN is always significantly slower than other methods.

In \cref{tab:extended-semantic-segmentation}, the metrics to measure predictive uncertainty show that VQ-BNN achieves the best performance in most cases. 
To be specific, we observe that the predictive performance of VQ-BNN in terms of NLL is comparable to that of BNN, and is always significantly better than that of DNN. Moreover, VQ-BNN is always better calibrated than other methods.

\Cref{fig:reliability-diagram-extended} is the reliability diagram with \emph{discrepancy}, the difference between confidence and accuracy. This figure is almost the same as \Cref{fig:reliability-diagram}, but it also represents the performance of VQ-DNN. This figure shows that VQ-DNN and VQ-BNN are better calibrated than DNN and BNN.

\paragraph{Qualitative results. }\label{app:qualitative-results}

\begin{figure*}
     \centering
     \begin{subfigure}[b]{0.20\textwidth}
         \centering
         \includegraphics[width=\textwidth,page=5]{resources/visualize/semantic-segmentation/visualize.pdf}
         \caption{Input image}
     \end{subfigure}
     \begin{subfigure}[b]{0.20\textwidth}
         \centering
         \includegraphics[width=\textwidth,page=6]{resources/visualize/semantic-segmentation/visualize.pdf}
         \caption{Ground truth}
     \end{subfigure}
     \vskip 0.1in

     \centering
     \begin{subfigure}[b]{0.20\textwidth}
         \centering
         \includegraphics[width=\textwidth,page=3]{resources/visualize/semantic-segmentation/visualize.pdf}
         \includegraphics[width=\textwidth,page=4]{resources/visualize/semantic-segmentation/visualize.pdf}
         \caption{DNN}
         \label{fig:visualize-semantic-segmentation:dnn}
     \end{subfigure}
     \begin{subfigure}[b]{0.20\textwidth}
         \centering
         \includegraphics[width=\textwidth,page=1]{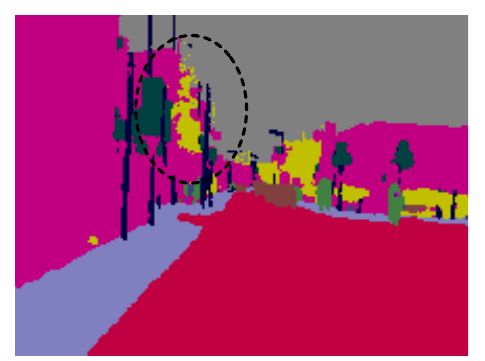}
         \includegraphics[width=\textwidth,page=2]{resources/visualize/semantic-segmentation/visualize.pdf}
         \caption{BNN}
         \label{fig:visualize-semantic-segmentation:bnn}
     \end{subfigure}
     \begin{subfigure}[b]{0.20\textwidth}
         \centering
         \includegraphics[width=\textwidth,page=9]{resources/visualize/semantic-segmentation/visualize.pdf}
         \includegraphics[width=\textwidth,page=10]{resources/visualize/semantic-segmentation/visualize.pdf}
         \caption{VQ-DNN}
         \label{fig:visualize-semantic-segmentation:vqdnn}
     \end{subfigure}
     \begin{subfigure}[b]{0.20\textwidth}
         \centering
         \includegraphics[width=\textwidth,page=7]{resources/visualize/semantic-segmentation/visualize.pdf}
         \includegraphics[width=\textwidth,page=8]{resources/visualize/semantic-segmentation/visualize.pdf}
         \caption{VQ-BNN}
         \label{fig:visualize-semantic-segmentation:vqbnn}
     \end{subfigure}
          
     \caption{
     \textbf{Qualitative results with semantic segmentation} for each method. 
     In \crefrange{fig:visualize-semantic-segmentation:dnn}{fig:visualize-semantic-segmentation:vqbnn}, the first row is the predictive result and the second row is the predictive confidence. A whiter background corresponds to higher confidence.}
     \label{fig:visualize-semantic-segmentation}
\end{figure*}

\begin{figure*}
     \centering
     \begin{subfigure}[b]{0.20\textwidth}
         \centering
         \includegraphics[width=\textwidth,page=1]{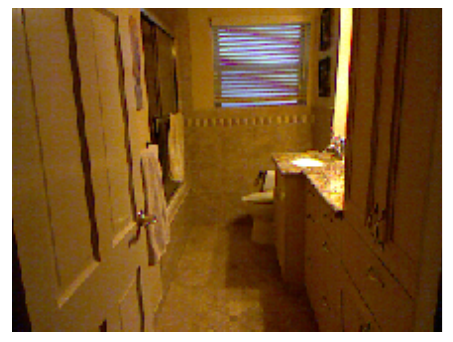}
         \caption{Input image}
     \end{subfigure}
     \begin{subfigure}[b]{0.20\textwidth}
         \centering
         \includegraphics[width=\textwidth,page=2]{resources/visualize/depth-estimation/visualize.pdf}
         \caption{Ground truth}
     \end{subfigure}
     \vskip 0.1in

     \centering
     \begin{subfigure}[b]{0.20\textwidth}
         \centering
         \includegraphics[width=\textwidth,page=3]{resources/visualize/depth-estimation/visualize.pdf}
         \includegraphics[width=\textwidth,page=4]{resources/visualize/depth-estimation/visualize.pdf}
         \caption{DNN}
         \label{fig:visualize-depth-estimation:dnn}
     \end{subfigure}
     \begin{subfigure}[b]{0.20\textwidth}
         \centering
         \includegraphics[width=\textwidth,page=5]{resources/visualize/depth-estimation/visualize.pdf}
         \includegraphics[width=\textwidth,page=6]{resources/visualize/depth-estimation/visualize.pdf}
         \caption{BNN}
         \label{fig:visualize-depth-estimation:bnn}
     \end{subfigure}
     \begin{subfigure}[b]{0.20\textwidth}
         \centering
         \includegraphics[width=\textwidth,page=7]{resources/visualize/depth-estimation/visualize.pdf}
         \includegraphics[width=\textwidth,page=8]{resources/visualize/depth-estimation/visualize.pdf}
         \caption{VQ-DNN}
         \label{fig:visualize-depth-estimation:vqdnn}
     \end{subfigure}
     \begin{subfigure}[b]{0.20\textwidth}
         \centering
         \includegraphics[width=\textwidth,page=9]{resources/visualize/depth-estimation/visualize.pdf}
         \includegraphics[width=\textwidth,page=10]{resources/visualize/depth-estimation/visualize.pdf}
         \caption{VQ-BNN}
         \label{fig:visualize-depth-estimation:vqbnn}
     \end{subfigure}
          
     \caption{
     \textbf{Qualitative results with depth estimation} for each method. 
     In \crefrange{fig:visualize-depth-estimation:dnn}{fig:visualize-depth-estimation:vqbnn}, the first row is the predictive result and the second row is the predictive confidence. For predictive result, a whiter area corresponds to deeper result. For predictive confidence, a whiter background corresponds to higher confidence.}
     \label{fig:visualize-depth-estimation}
\end{figure*}

\Cref{fig:visualize-semantic-segmentation} shows the qualitative comparison of the predictions for each method. In this figure, DNN is overconfident, i.e., confidence is high even when the prediction is wrong, and uncertainty is mostly distributed at the edge of the objects. 
The uncertainty of BNN is distributed on the edge as in the case of the DNN, but is also distributed in the misclassified areas. 
VQ-DNN is less overconfident compared to DNN. Also, although VQ-DNN does not identify all of the misclassifications compared to BNN, it sometimes estimates high uncertainty in the misclassified areas. The uncertainty of VQ-BNN is similar to the BNN as we expected.
This figure is a comparison on a static image sequence; see \Cref{fig:qualitative-analysis} for a comparison on a dynamic image sequence.

\paragraph{Edge predictive performance.}

In \cref{sec:semantic-segmentation}, we argue that misclassification in semantic segmentation frequently occurs at the edge of the object, and VQ-BNN shows high predictive performance at the edge of the object. To demonstrate these, we propose edge predictive performance, which is the predictive performance for the pixel of the object edge. We use the Sobel operator to detect the object edge and find a pixel that satisfies the following conditions:
\begin{align}
	\sqrt{G_{\bm{x}}^{2} + G_{\bm{y}}^{2}} \geq \Gamma
\end{align}
where $G_{\bm{x}}$ and $G_{\bm{y}}$ are the pixel-wise results of the Sobel operator of $\bm{x}$- and $\bm{y}$-coordinate for input image, respectively, and $\Gamma$ is non-negative threshold. For example, if $\Gamma$ is 0, all pixels will be edge pixels. We obtain the edge predictive performance by measuring the performance for the pixel corresponding to this condition.

\Cref{tab:edge-performance} shows the edge predictive performance of each method when thresholds are 0.0, 0.5, and 1.0 in the case of \textsc{CamVid \& U-Net}. The results of this table support our claims. See \cref{sec:semantic-segmentation} for analysis on this table.

\paragraph{Temporal consistency. }

As described in \cref{sec:vqbnn}, VQ-BNN relies on temporal consistency of data streams. To evaluate the sensitivity to the temporal consistency, we conduct an experiment of reducing frame rate, which decreases temporal consistency. In this experiment, we subsample 30fps CamVid image sequences every 2 to 10 frames; then, we evaluate VQ-BNN on these 3 to 15 fps video streams.

\begin{figure}
\centering
\begin{subfigure}[b]{0.30\textwidth}
\centering
\includegraphics[width=\textwidth]{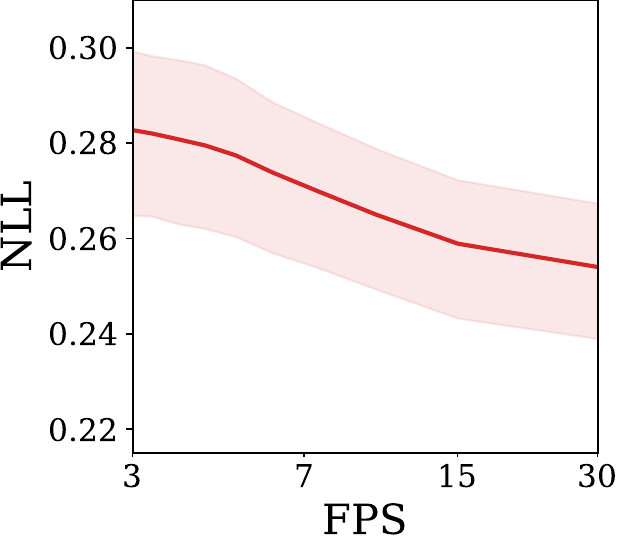}
\label{}
\end{subfigure}
\vskip -0.15in
	
\caption{
\textbf{NLL of VQ-BNN for the video frame rate (FPS)}.
}
\label{fig:fps}
\end{figure}

\Cref{fig:fps} shows the NLL of VQ-BNN for the video frame rates. According to the result, when the frame rate is reduced, the predictive performance in terms of NLL is degraded. For the frame rate is 10fps or higher, VQ-BNN significantly improves the predictive performance, whereas for the frame rate is 5fps or less, it provides only a marginal performance improvement.

\paragraph{Improving performance by using future predictions. }

\cref{sec:vqbnn} discusses that VQ-BNN improves predictive performance by using past predictions. Similarly, VQ-BNN allows us to improve predictive performance by using not only past predictions but also \emph{future} predictions. For example, for a classification task such as semantic segmentation, \cref{eq:vqbnn-classification} is modified as follows:
\begin{align}
	p(\outs \vert \strs, \mathcal{D}) \simeq \sum_{t=J}^{-K} \pi(\dats_{t} \vert \strs) \, \texttt{Softmax}(\nn (\dats_{t}, \lats_{t}))
	\label{eq:vqbnn-future-frame}
\end{align}
where $J$ is the number of future predictions. In other words, the predictive distribution is the sum of one prediction for the input data, $K$ past predictions, and $J$ future predictions.

\begin{figure}
\centering
\begin{subfigure}[b]{0.30\textwidth}
\centering
\includegraphics[width=\textwidth]{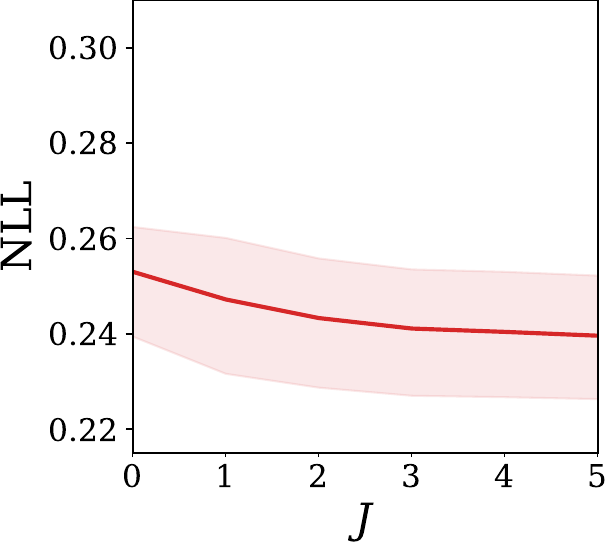}
\label{}
\end{subfigure}
\vskip -0.15in
	
\caption{
\textbf{NLL of VQ-BNN for the number of future frames ($J$)}.
}
\label{fig:future-frame}
\end{figure}

\Cref{fig:future-frame} shows the predictive performance of VQ-BNN in terms of NLL for the number of future frames $J$ with semantic segmentation on the CamVid. In this experiment, VQ-BNN uses 5 past predictions. According to the result, as $J$ increases until 5, the predictive performance improves.

In stream processing, VQ-BNN cannot estimate predictions for future inputs that are not given. Instead, VQ-BNN calculates the predictive distribution by using \cref{eq:vqbnn-future-frame} after $J$ inputs are given. As a result, future predictions improves predictive performance, but increases latency. For example, if VQ-BNN uses one future frame to process 30fps video, the latency increases by 1/30 second. For this reason, we recommend using one or two future predictions in a practical situation.

\subsection{Depth Estimation}\label{sec:extended-depth-estimation}

\cref{sec:depth-estimation} shows the performance of VQ-BNN with depth estimation. This section uses SegNet additionally, and provides detailed information and experimental results on this experiment.

\paragraph{Computational and predictive performances. }

\Cref{tab:extended-depth-estimation} shows the computational and predictive performance of various methods. We measure \textsc{Bat Thr} with batch size of 10. For other setups, we follow \cref{sec:extended-semantic-segmentation}.

We use additional metrics commonly used for depth estimation to measure predictive accuracy:  mean relative error (\textsc{Rel}, $\downarrow$) and thresholded accuracies ($\delta$, $\uparrow$). 
In the case of \textsc{NYUDv2 \& U-Net}, all these metrics show that the predictive performance of VQ-BNN is the best. 
In the case of \textsc{NYUDv2 \& SegNet}, the predictive performance of VQ-DNN and VQ-BNN are better then that of DNN and BNN, respectively. However, it might be an unusual result that VQ-DNN is better than VQ-BNN. The reason for the result is that DNN is better than BNN in this case.

\begin{table*}[t]
\vskip 0.1in
\begin{center}
\begin{small}
\begin{sc}

  \begin{tabular}{lcccccccccc}
    \toprule

    Method & \thead{Bat Thr\\(Img/Sec)} & \thead{Str Thr\\(Img/Sec)} & NLL & \thead{Acc\\(\%)} & \thead{Acc$_{90}$\\(\%)} & \thead{Unc$_{90}$\\(\%)} & \thead{IoU\\(\%)} & \thead{IoU$_{90}$\\(\%)} & \thead{Freq$_{90}$\\(\%)} & \thead{ECE\\(\%)} \\
    \midrule
	DNN			
	& \textbf{30.7} & \textbf{24.1}
	& 0.787 & 78.2 & 90.4 & 73.4 & 47.9 & 63.6 & 60.7 & 8.00 \\
	BNN			
	& 2.42 & 0.833 
	& \textbf{0.509} & \textbf{83.6} & \textbf{94.9} & \textbf{81.4} & \textbf{51.8} & \textbf{72.5} & 59.6 & \textbf{2.00} \\
	VQ-BNN		
	& 30.4 & 23.7 
	& 0.552 & 82.4 & 93.7 & 78.0 & 49.6 & 68.5 & 61.6 & 4.12 \\
	AVP			
	& 10.5 & 8.31 
	& 0.894 & 78.2 & 88.4 & 62.6 & 47.9 & 60.3 & \textbf{70.2} & 11.5 \\

    \bottomrule
  \end{tabular}

\end{sc}
\end{small}
\end{center}
\caption{
\textbf{Computational and predictive performance with semantic segmentation for sampling-free BNN (AVP).}
}
\label{tab:semantic-segmentation-performance-for-sampling-free}

\vskip -0.1in
\end{table*}

\paragraph{Qualitative results. }

\Cref{fig:visualize-depth-estimation} shows the qualitative comparison of the predictions for each method. DNN predicts uniform uncertainty in the regression task; the uncertainty estimated by DNN is meaningless. BNN estimates high uncertainty at the misestimation areas. In this example, NNs incorrectly estimate the depth of the corridor window, and BNN estimates high uncertainty at the window area. VQ-DNN does not exactly find out which results are incorrectly estimated, but exhibits high uncertainty at the object edge. Since it is difficult to estimate the depth of object edges, the predictive performance of VQ-DNN is better than that of DNN. VQ-BNN represents both types of the uncertainties, so the uncertainty of VQ-BNN is reliable.

\section{Comparison with Sampling-free BNN}\label{sec:sampling-free}

As mentioned in \cref{sec:related-work}, most sampling-free BNNs only work well for shallow NNs with several layers or less. \cite{postels2019sampling} concurrently proposed Approximated Variance Propagation (AVP), and, to the best of our knowledge, it is the only sampling-free BNN with enough layers for practical tasks such as semantic segmentation. AVP transforms the trained BNN with MC dropout into an approximate neural network that predicts a Gaussian distribution with a diagonal covariance matrix. This section compares the performance of VQ-BNN and AVP. 

In principle, if our understanding is correct, AVP is not designed for classification tasks such as semantic segmentation, and \cite{postels2019sampling} did not provide a quantitative comparison with semantic segmentation experiment in the paper. Neural networks for semantic segmentation have to predict categorical distributions. However, the neural network transformed by AVP predicts the means and standard deviations of the Gaussian distribution, and standard deviations are not applicable to categorical distribution.

To solve this problem, we use a method to convert the standard deviations of the Gaussian distribution to the confidences of the categorical distribution.  
We follow the \cite{postels2019sampling} and derive the standard deviation of BNN's predictions. Subsequently, we generate a one-to-one correspondence between standard deviation and confidence based on the original BNN predictions. We only use Bayesian SegNet \cite{kendall2015bayesian} in this experiment because \cite{postels2019sampling} does not provide a transformation for skip layers. We follow \cite{postels2019sampling} for other experimental setups.

\Cref{tab:semantic-segmentation-performance-for-sampling-free} shows the quantitative comparison of DNN, BNN, VQ-BNN, and AVP, on the CamVid dataset. 
The results shows that the predictive result of AVP is more inaccurate than that of BNN and VQ-BNN, and is comparable to that of DNN. Moreover, AVP shows less accurate estimation of uncertainty than BNN and VQ-BNN.
The throughput of AVP is 34$\%$ of that of DNN while the throughput of VQ-BNN is 98$\%$.

In conclusion, AVP has the following limitations: 
First, as \cite{postels2019sampling} mentioned in the paper, \emph{``the magnitude of the approximated uncertainty (by AVP) is much lower than the sampling-based uncertainty''}, i.e., AVP is not well-calibrated. To achieve high predictive performance, it seems essential to calibrate the results of AVP at a granular level.
Second, AVP can transform a limited kind of neural network. AVP only deals with BNNs that contain noise injection layers such as MC dropout and layers for which the transformation rules are known. 
Third, AVP predicts only Gaussian distribution with diagonal covariance matrix. As in this section, AVP requires additional methods to convert standard deviations to apt uncertainty to solve discrete problems such as classification tasks. 
Finally, AVP is considerably slower than VQ-BNN because AVP predicts mean and standard variance separately.

\end{document}